%% file: boxcomm.tex
\documentclass[letterpaper]{article} % DO NOT CHANGE THIS
\usepackage[preprint]{aaai2027} % DO NOT CHANGE THIS
\usepackage[hyphens]{url} % DO NOT CHANGE THIS
\usepackage{graphicx} % DO NOT CHANGE THIS
\usepackage{natbib} % DO NOT CHANGE THIS AND DO NOT ADD OPTIONS
\usepackage{caption} % DO NOT CHANGE THIS AND DO NOT ADD OPTIONS
\usepackage{booktabs}
\usepackage{multirow}
\usepackage{amsmath}
\usepackage{pifont}
\newcommand{\cmark}{\ding{51}}
\newcommand{\xmark}{\ding{55}}
\usepackage{array}
\usepackage{amssymb}
\usepackage{needspace}
\usepackage{listings}
\lstdefinestyle{boxcommprompt}{
  basicstyle=\ttfamily\scriptsize,
  breaklines=true,
  columns=fullflexible,
  keepspaces=true,
  showstringspaces=false,
  xleftmargin=0.8em,
  xrightmargin=0.4em,
  aboveskip=0.35em,
  belowskip=0.55em
}

\begin{document}

\title{BoxComm: Benchmarking Category-Aware Commentary Generation and Rhythm Planning in Boxing}

\author{%
  Kaiwen Wang\textsuperscript{1}\equalcontrib,
  Kaili Zheng\textsuperscript{1}\equalcontrib,
  Rongrong Deng\textsuperscript{2},
  Eric Shaobo Wang\textsuperscript{3},\\
  Yiming Shi\textsuperscript{1},
  Yang Qingyuan\textsuperscript{4},
  Chenyi Guo\textsuperscript{1}\corresponding,
  Ji Wu\textsuperscript{1}\corresponding}
\affiliations{%
  \textsuperscript{1}Tsinghua University, Beijing, China \quad
  \textsuperscript{2}Beijing Sport University, Beijing, China\\
  \textsuperscript{3}University of Wisconsin--Madison\\
  \textsuperscript{4}Xinjiang Region Sports Science Research Center\\
  \{wkw23,zkl25\}\url{@mails.tsinghua.edu.cn} \quad \url{guochy@mail.tsinghua.edu.cn} \quad \url{wuji_ee@tsinghua.edu.cn}}
% Preserve the user-requested first-page, double-column teaser layout.
\makeatletter
\g@addto@macro\@maketitle{%
  \begin{center}
    \includegraphics[width=\textwidth]{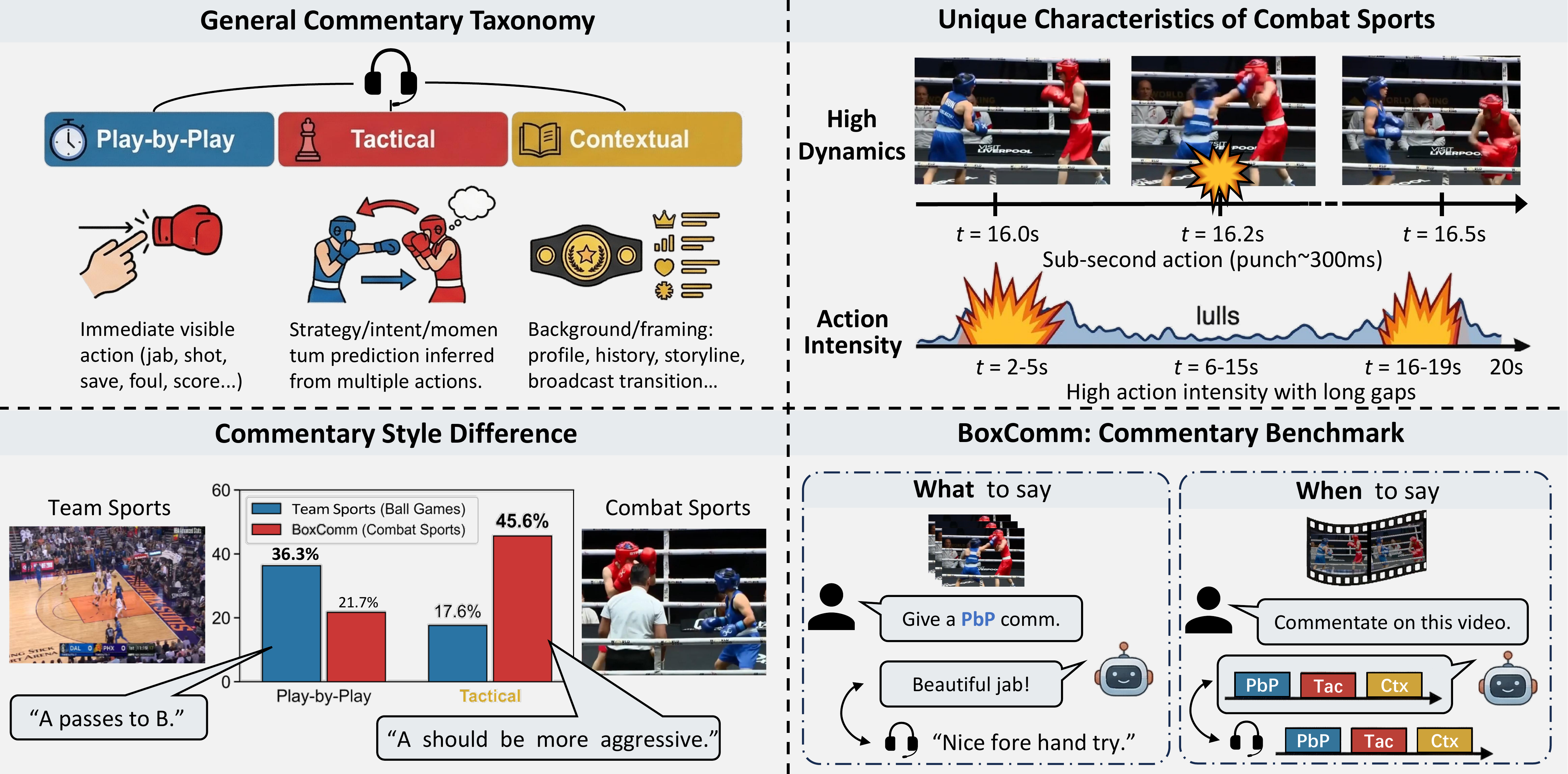}
    \captionof{figure}{Overview of commentary categories, boxing-specific challenges,
    and the BoxComm benchmark.}
    \label{fig:teaser}
  \end{center}
}
\makeatother
\maketitle


\begin{abstract}
Recent multimodal large language models (MLLMs) have spurred interest in sports commentary generation. Yet professional commentary goes beyond describing visible actions: a system must also interpret tactics, provide context, and decide when to speak. Boxing is especially challenging because rapid actions and exchanges demand both immediate description and tactical interpretation. We therefore organize sports commentary into three functional categories: play-by-play, tactical, and contextual. Based on this taxonomy, we introduce \textbf{BoxComm}, the first category-aware boxing commentary benchmark. It contains 52K temporally aligned professional sentences from 445 World Boxing Championship broadcasts and 260K structured punch events. These data support two evaluations: category-conditioned generation measures semantic agreement with professional reference commentary under a specified category, while rhythm assessment compares silence and category timing over full matches. Our evaluation of state-of-the-art MLLMs shows substantial gaps in both content consistency and temporal rhythm. We also introduce a two-stage system: Event-Informed Commentary Generation (EIC-Gen) augments category-conditioned generation with recent punch-event cues, while a Commentary Planner predicts silence or a category from match state and past events. Fine-tuning Qwen3-VL with EIC-Gen raises content consistency from 24.1\% to 29.1\%, including gains of 7.9 points for play-by-play and 9.0 for tactical commentary. Coupled with the fine-tuned generator, the Planner achieves 0.141 mean t-IoU and 0.187 four-state JS, outperforming the evaluated streaming baselines. Commentary and event annotations are publicly available.
\end{abstract}
\begin{links}
    \link{Data}{https://huggingface.co/datasets/gouba2333/BoxComm-Dataset}
\end{links}



\section{Introduction}

Recent multimodal large language models (MLLMs) have made substantial progress in video understanding and generation \cite{hurst2024gpt,wang2025internvl3,lin2024video,an2025llava}. Sports commentary is a demanding application: a system must understand what is happening and why it matters, then decide when to communicate that information during a continuous broadcast. Such systems could improve accessibility, support broadcast production, and enable interactive fan experiences \cite{xia2024language,andrews2024designing,yan2026proact}.

Several datasets and benchmarks support sports commentary research. SoccerNet-Caption \cite{mkhallati2023soccernet} and MatchTime \cite{rao2024matchtime} provide large-scale soccer commentary data, while BH-Commentary \cite{zhang2024descriptive} focuses on basketball. Subsequent studies \cite{li2025multi,you2025timesoccer,rao2025towards} show that domain-specific fine-tuning can improve commentary generation. However, these team-sports benchmarks focus on annotated moments or highlight clips, without separating category-conditioned content from speaking rhythm over continuous broadcasts. The temporal and narrative structure of combat sports remains largely unexamined.

Boxing, in particular, presents distinct challenges in two key respects. First, boxing actions unfold within hundreds of milliseconds. The visual difference between an effective punch and a blocked or glancing attempt can be subtle, but it changes how the action should be narrated. By comparison, key events in team sports, such as goals, passes, and fouls, are usually more visible and farther apart. Second, rapid exchanges alternate with quieter intervals. Rather than filling the broadcast with continuous action description, commentators use these intervals to interpret tactics, technique, and matchup dynamics. Following prior linguistic analysis of sports-announcer talk \cite{ferguson1983sports}, we organize professional boxing commentary into three functional categories: play-by-play (real-time description of ongoing action), tactical (analysis of strategy, technique, and matchup dynamics), and contextual (background information such as fighter records, historical significance, and crowd reactions). Compared with StreamingVLM's team-sports sample \cite{xu2025streamingvlm}, boxing devotes more time to tactical commentary (45.6\% vs 17.6\%) and less to play-by-play (21.7\% vs 36.3\%) (Figure~\ref{fig:teaser}). This contrast motivates category-aware evaluation of both immediate exchanges and the tactical narrative developed around them.

To fill this gap, we introduce BoxComm, comprising 445 World Boxing Championship matches and 52K temporally aligned professional commentary sentences labeled as play-by-play, tactical, or contextual. These sentence-level functional labels establish BoxComm as the first category-aware boxing commentary benchmark. We define two complementary evaluations. Category-conditioned generation supplies a video segment, preceding commentary, and a target type, then tests whether the output preserves the ground-truth commentary's core meaning. Commentary rhythm assessment supplies neither timing nor category labels: over a full match, the system decides when to speak and which category to use. Its output is a timeline of silence, play-by-play, tactical, and contextual states, whose duration distribution is compared with ground truth. The protocols separate category-conditioned \emph{what to say} from autonomous \emph{when to say it}. This setting extends recent proactive video generation by adding explicit commentary categories to continuous boxing broadcasts \cite{yan2026proact,afzal2026real}. Because more than one commentary choice may be valid, we also conduct an anonymous window-level human evaluation without reference matching.

We evaluate a range of state-of-the-art MLLMs on both evaluations and find persistent limitations in boxing commentary generation. Even after task-specific fine-tuning, models struggle to describe fast boxing actions and maintain appropriate rhythm. The results show that current models still lack fine-grained perception and domain reasoning. We therefore propose Event-Informed Commentary Generation (EIC-Gen), which gives MLLMs concise cues from a dedicated punch-event extractor. Combined with recent commentary history, these cues improve both play-by-play and tactical generation after fine-tuning. For autonomous streaming, our Commentary Planner predicts silence or a category from match state and past events, improving category timing and four-state rhythm. Human raters also find the resulting system more grounded and prefer it to LiveCC and Proact-VL for timing and overall quality.

Our contributions are summarized as follows:

\begin{itemize}
    \item We introduce BoxComm, a large-scale boxing commentary dataset with 445 World Boxing Championship match videos and 52K sentence-level annotations. Its three-way labels establish BoxComm as the first category-aware benchmark for boxing commentary.
    \item We introduce complementary protocols for category-conditioned content generation and autonomous four-state commentary rhythm. Our MLLM benchmark reveals clear limitations in content consistency and commentary timing.
    \item We propose EIC-Gen for content generation and an Commentary Planner for autonomous streaming. Event cues improve play-by-play and tactical consistency; the Planner improves category timing and four-state rhythm over the evaluated baselines.
\end{itemize}

\section{Related Work}

\subsection{MLLMs for Video Understanding}

Video MLLMs use sampled visual sequences as input to a language model. Video-LLaMA and Video-ChatGPT introduced video-conditioned dialogue \cite{zhang2023video,maaz2024video}, while GPT-4o, Gemini, InternVL3.5, LLaVA-Video, Qwen3-VL, and Video-LLaVA extend these capabilities to broader multimodal and temporal tasks \cite{hurst2024gpt,team2023gemini,wang2025internvl3,zhang2024llava,bai2025qwen3,lin2024video}. Benchmarks such as Video-MME test temporal grounding and dense understanding \cite{fu2025video}. Most of these systems answer prompted questions or caption bounded clips. Commentary adds another challenge: the system must identify actions and decide which events are worth describing as the video unfolds.

\subsection{Sports Video Analysis and Commentary}
Sports video research spans action spotting and player analysis in SoccerNet, NSVA, and TenniSet \cite{giancola2018soccernet,deliege2021soccernet,wu2022sports,faulkner2017tenniset}. SPORTU and SportR expose expert-level gaps in sports-specific MLLM reasoning \cite{xia2024sportu,xia2025sportr}, while FineBadminton connects fine-grained actions to tactics \cite{he2025finebadminton}. These works show the need to connect fine-grained perception with professional sports narration.

Commentary generation primarily covers soccer \cite{mkhallati2023soccernet,qi2023goal,rao2024matchtime,you2025timesoccer,rao2025towards,li2025multi}, basketball highlights \cite{zhang2024descriptive}, and tennis \cite{yan2016generating,liu2026tennisexpert}. Earlier systems verbalize recorded events, whereas recent MLLMs learn scene-to-sentence generation. SCBench evaluates six sports \cite{ge2024scbench}, but excludes combat sports and does not explicitly evaluate commentary functions.

Existing evaluations remain clip- or timestamp-based: SoccerNet-Caption combines captioning and temporal metrics, MatchTime adds GPT-based semantic and professional criteria, and SCBench judges short clips \cite{mkhallati2023soccernet,rao2024matchtime,ge2024scbench}. \citet{mori-etal-2025-live} generate play-by-play and retrieval-grounded background for football. BoxComm also evaluates tactical commentary and silence-aware rhythm over full boxing matches. Combat-sports work, meanwhile, remains centered on action recognition without professional commentary modeling \cite{kumar2025boxingvi,sahoo2024boxmac,wang2026boxmind}. This leaves professional combat-sports narration largely unexplored.

\subsection{Streaming and Proactive Generation}
StreamingVLM and LiveCC generate language over continuous video \cite{xu2025streamingvlm,chen2025livecc}; StreamBridge adds memory and proactive activation \cite{wang2025streambridge}; and MMDuet2 and Proact-VL model whether and when to respond \cite{wang2025mmduet2,yan2026proact}. Pause-Aware Decoding separates \emph{what to say} from \emph{when to say it} for game commentary \cite{afzal2026real}. These works mainly treat proactivity as response activation or pause control. BoxComm also asks which professional commentary function to use whenever the system speaks, combining silence-aware timing with category-conditioned content.

\section{BoxComm Dataset}

\subsection{Dataset Collection}
The BoxComm dataset contains 445 live broadcasts from the 2025 World Boxing Championships. These matches represent top-level competitive boxing and cover a wide range of weight classes and tactical styles. As we study continuous commentary rather than isolated highlights, we retain full broadcasts, including active exchanges, quiet periods, and between-round transitions. All videos are at least 25 frames per second and 720p resolution, preserving rapid, subtle actions, including jabs lasting about 200 milliseconds, for annotation and full-match evaluation.

\begin{figure}[htbp]
  \includegraphics[width=\linewidth]{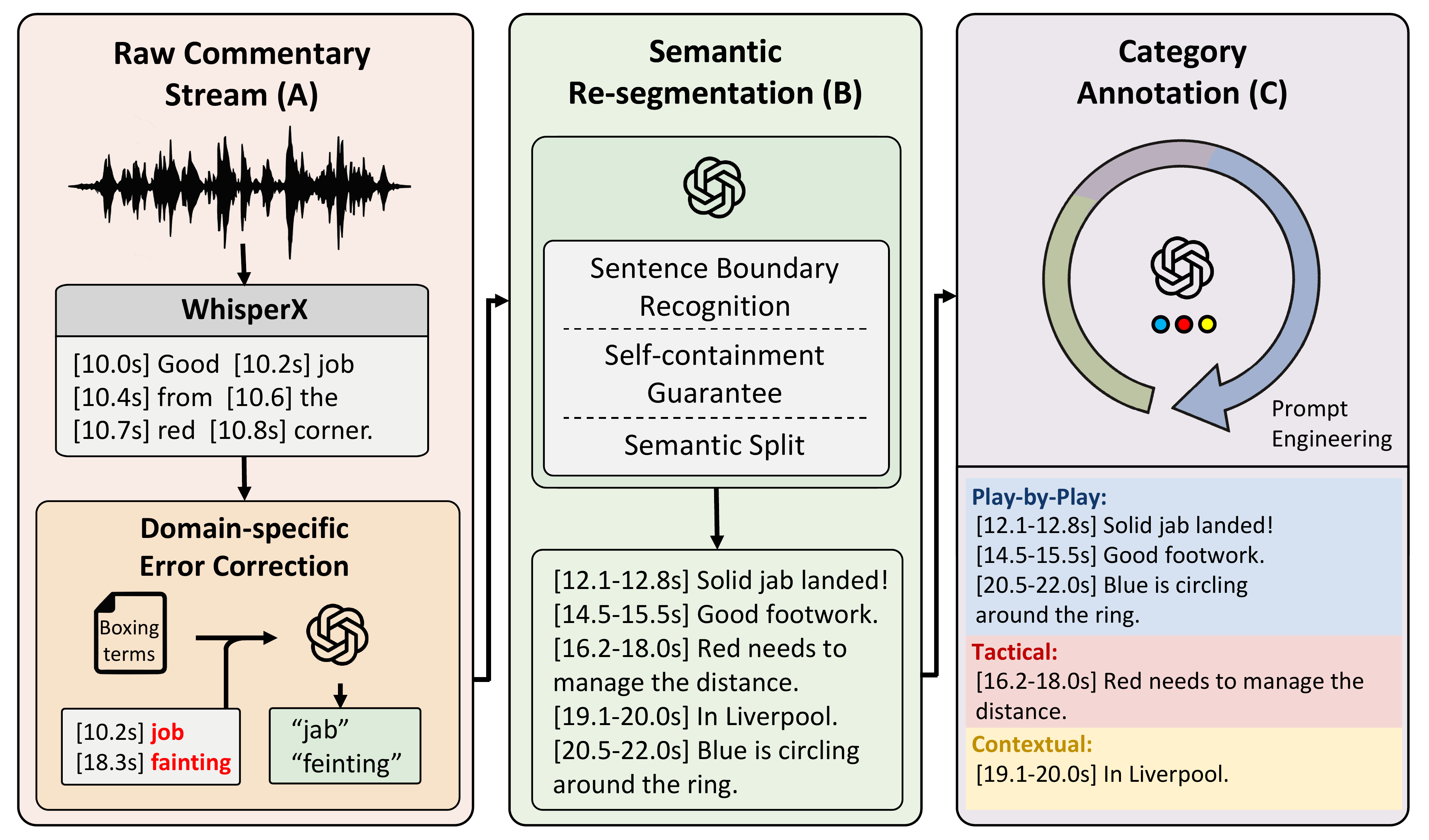}
  \caption{The pipeline for commentary extraction, semantic re-segmentation, and category annotation.}
  \label{fig:asr}
\end{figure}

\begin{figure}[!b]
    \centering
    \includegraphics[width=\linewidth]{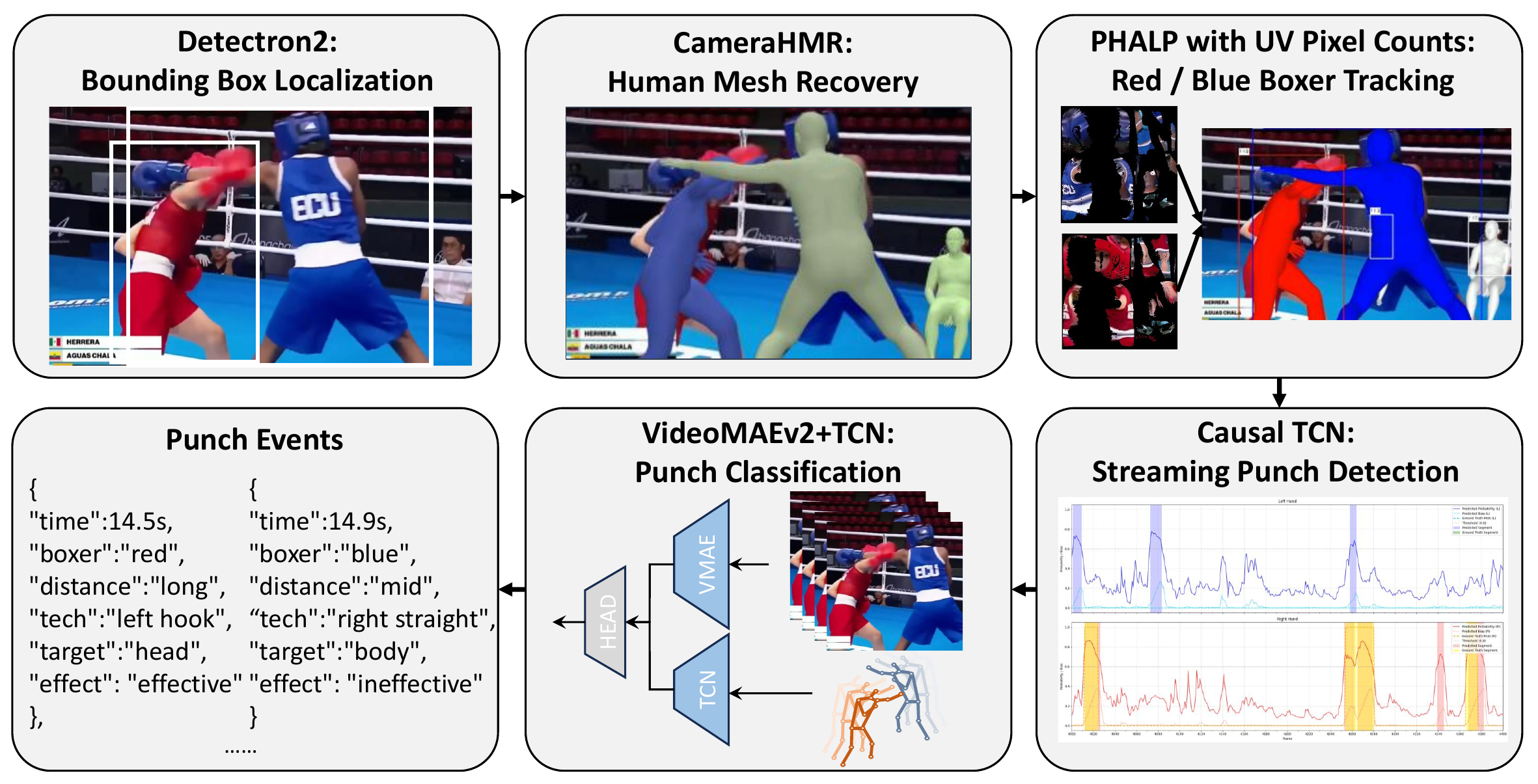}
    \caption{Punch event extraction pipeline.}
    \label{fig:BoxMind}
\end{figure}

\begin{figure*}[!t]
    \centering
    \includegraphics[width=\textwidth]{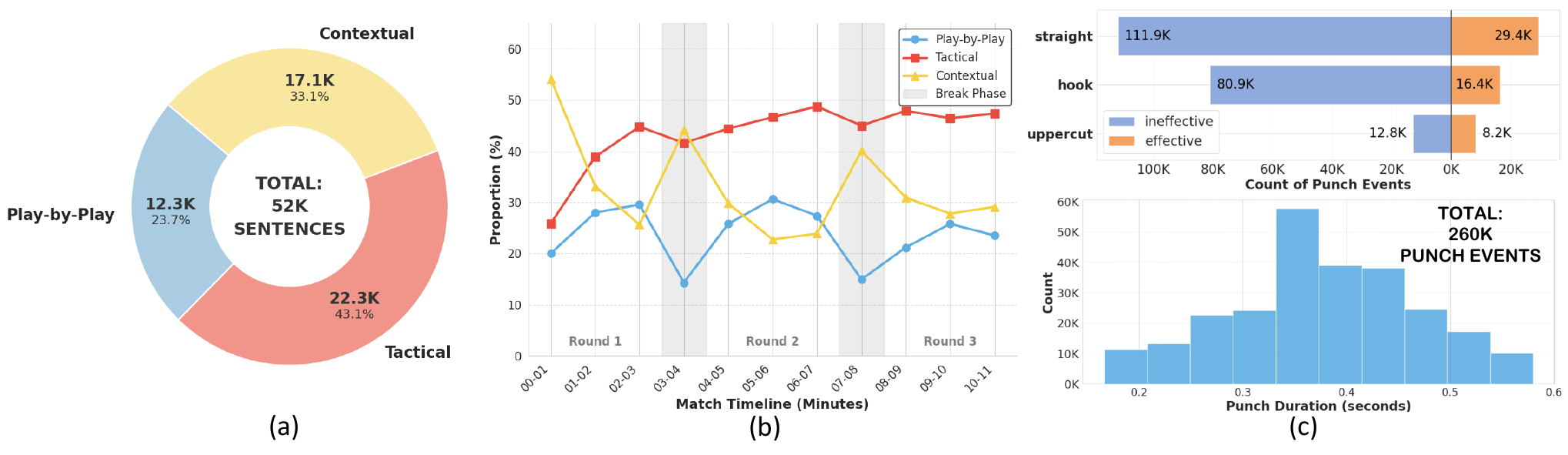}
    \caption{BoxComm dataset statistics. (a) Sentence-count category proportions. (b) Temporal category distribution across the match. (c) Fine-grained punch event attributes.}
    \label{fig:stat}
\end{figure*}

We first use WhisperX to transcribe the broadcast audio with word-level timestamps (Figure~\ref{fig:asr}). GPT-5.2 then corrects domain-specific terms and segments the transcript into complete utterances at semantic boundaries and natural speech pauses. Brief Play-by-Play descriptions are kept separate from neighboring Tactical analysis so that short action descriptions remain closely aligned with rapidly changing boxing actions.

\subsection{Category Annotation}
Each segmented sentence in BoxComm is assigned to one of three categories: Play-by-Play, Tactical, or Contextual. We inspect borderline cases to refine the category boundaries and classification prompt, after which GPT-5.2 assigns labels using the revised definitions and neighboring context. The final dataset contains 52K temporally aligned, categorized sentences for studying professional boxing commentary across full broadcasts.

\subsection{Punch Event Extraction Pipeline}
To capture rapid, subtle boxing actions, we follow the fine-grained framework of BoxMind \cite{wang2026boxmind} (Figure~\ref{fig:BoxMind}). Detectron2 \cite{detectron2} and CameraHMR \cite{camerahmr} localize the boxers and recover their meshes, while PHALP \cite{4dhumans} maintains red/blue identities. A causal TCN detects punch candidates online; once a candidate's end frame is observed, a fused VideoMAEv2--TCN model \cite{wang2023videomae,lea2017temporal} estimates its attributes from the completed segment. We train these modules on BoxMind, an open-source boxing dataset \cite{olympic-boxing-punch-dataset-2021}, and a proprietary punch dataset, then apply them to all 445 matches. Each completed event records its time, boxer, distance, technique, target, and effectiveness.

\subsection{Dataset Statistics}

Figure \ref{fig:stat} summarizes the BoxComm dataset. It contains 52K commentary sentences, including 22.3K Tactical sentences. Category use also changes over the course of a match. Tactical commentary remains common from the latter part of the first round onward, while Contextual commentary appears most often at the start of a match and between rounds. BoxComm also contains 260K fine-grained punch events from the extraction pipeline. These events record punch type and effectiveness, and most last $0.3$--$0.5$ seconds.

To establish a standard benchmark, we partition the 445 matches chronologically: the first 405 form BoxComm-Train and the final 40 form BoxComm-Eval. This split targets generalization to later events rather than a new broadcast domain: both sets use the same broadcast-source family, while BoxComm-Eval covers all 20 weight classes (10 women's and 10 men's). It contains 5.6K commentary sentences and 23.4K fine-grained punch events in total.

\section{Benchmark Protocols}

\begin{figure*}[!t]
    \centering
    \includegraphics[width=0.9\textwidth]{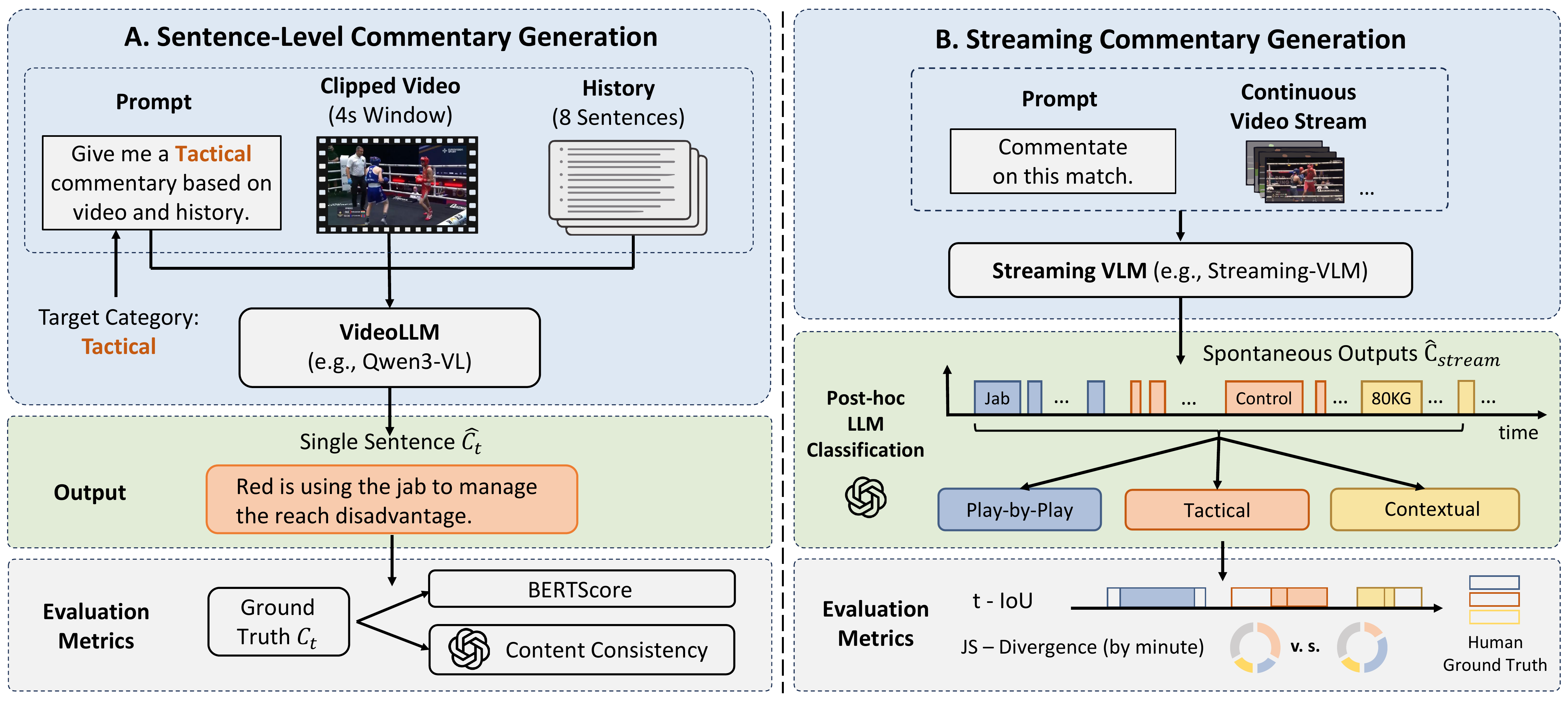}
    \caption{Evaluation protocols: category-conditioned commentary generation and streaming commentary rhythm assessment.}
    \label{fig:architecture}
\end{figure*}

We evaluate two complementary capabilities (Figure~\ref{fig:architecture}): Category-Conditioned Generation tests \emph{what to say} when a commentary type is specified, while Commentary Rhythm Assessment tests \emph{when to speak} and which type to generate over continuous video.

\subsection{Category-Conditioned Generation}

Category-Conditioned Generation tests sentence content under a given commentary type. For a target sentence beginning at $t$, the model receives the preceding video window $V_{[t-k,t]}$ with $k_v=4$ seconds and the eight latest commentary sentences ending before $t$. It is also given a target category $c \in \{\text{play-by-play}, \text{tactical}, \text{contextual}\}$. The model then generates one sentence $\hat{C}_t$ that should preserve the ground-truth commentary's core meaning in the requested form. Because the category is given, errors mainly reflect sentence content rather than category selection. Reporting each category separately shows how well a model grounds immediate action, tactical interpretation, and background context.

Professional sports commentary can be phrased in many ways, so lexical metrics often penalize valid paraphrases. We use BERTScore to measure semantic similarity, but similarity alone does not establish that a generation describes the same boxing fact. We therefore add Content Consistency: GPT-5.2 judges whether the generated sentence describes the same action or topic as the category-specific reference without contradicting its core meaning.

\subsection{Commentary Rhythm Assessment}

Commentary Rhythm Assessment removes ground-truth timing and category prompts. At each time step, the system decides whether to remain silent or produce one of the three commentary types. The metrics evaluate category-aware interval alignment and duration allocation rather than content correctness. Treating Silence as an explicit state makes speaking frequency part of the task rather than an unscored absence. The resulting four-state timeline captures both speaking time decisions and narrative function.

For autonomous systems, we retain each system's sentence onsets and convert its text into finite-rate, non-overlapping speech intervals. Only the text that can be spoken before the next onset or video boundary is retained, and GPT-5.2 classifies each resulting sentence as play-by-play, tactical, or contextual. The GT Ref.\ row retains its original reconstructed intervals and full-sentence labels (appendix). We assess the resulting timelines with two metrics:

\textbf{Temporal Intersection-over-Union (t-IoU)} measures alignment between predicted and reference commentary intervals. For each category $c$, we compute a symmetric sentence-level t-IoU:
\begin{equation}
\begin{aligned}
t\text{-}IoU_c=\frac{1}{2}\Bigg(&\frac{1}{|P_c|}\sum_{p\in P_c}\max_{g\in G_c}\operatorname{IoU}(p,g)\\
&+\frac{1}{|G_c|}\sum_{g\in G_c}\max_{p\in P_c}\operatorname{IoU}(g,p)\Bigg)
\end{aligned}
\end{equation}
where $P_c$ and $G_c$ are the predicted and ground-truth intervals for category $c$. We report the mean over all three commentary categories.

\textbf{Silence-Aware Four-State Jensen--Shannon Divergence (JS)} measures agreement in duration allocation on the absolute timeline. For each valid video-minute cell $m$, we form predicted $Q_m$ and ground-truth $P_m$ duration distributions over $\mathcal{S}=\{\text{silence},\text{play-by-play},\text{tactical},\text{contextual}\}$ after deterministic latest-onset state assignment, let $M_m=(P_m+Q_m)/2$, and compute
\begin{equation}
\begin{aligned}
\operatorname{JS}(P_m,Q_m)
&=H_2(M_m)-\tfrac{1}{2}H_2(P_m)-\tfrac{1}{2}H_2(Q_m),\\
H_2(R)&=-\sum_{s\in\mathcal{S}}R(s)\log_2 R(s).
\end{aligned}
\end{equation}
Using base-2 entropy bounds JS in $[0,1]$. We average over valid video-minute cells; lower is better. This metric measures four-state duration occupancy, including silence, but not semantic accuracy, temporal order, or repetition within the generated stream.

Together, these metrics evaluate category-aware temporal rhythm in continuous commentary.

\begin{table*}[!t]
\centering
\small
\setlength{\tabcolsep}{3.5pt}
\begin{tabular}{l|c|cc|cc|cc|cc}
\toprule
\multirow{2}{*}{\textbf{Model}} & \multirow{2}{*}{\textbf{EIC-Gen}} & \multicolumn{2}{c|}{\textbf{Play-by-Play}} & \multicolumn{2}{c|}{\textbf{Tactical}} & \multicolumn{2}{c|}{\textbf{Contextual}} & \multicolumn{2}{c}{\textbf{Overall}} \\
\cmidrule{3-10}
 & & \textbf{BERTScore} & \textbf{Acc (\%)} & \textbf{BERTScore} & \textbf{Acc (\%)} & \textbf{BERTScore} & \textbf{Acc (\%)} & \textbf{BERTScore} & \textbf{Acc (\%)} \\
\midrule
GPT-4o-mini & \xmark & 85.9 & 15.2 & 85.7 & 18.5 & 86.1 & 14.2 & 85.9 & 16.4 \\
GPT-4o-mini & \cmark & 85.6 & 18.5 & 85.3 & 22.2 & 85.9 & 16.7 & 85.6 & 19.5 \\
\midrule
LLaVA-OV1.5-7B & \xmark & 87.1 & 11.2 & 86.7 & 11.9 & 86.8 & 9.4 & 86.8 & 10.9 \\
LLaVA-OV1.5-7B & \cmark & 86.5 & 13.3 & 85.8 & 24.8 & 85.7 & 21.3 & 85.9 & 21.3 \\
InternVL3.5-8B & \xmark & 84.8 & 23.9 & 84.6 & 32.1 & 84.8 & 15.5 & 84.7 & 24.6 \\
InternVL3.5-8B & \cmark & 84.8 & 21.3 & 84.7 & 27.8 & 84.9 & 16.9 & 84.8 & 22.7 \\
Qwen2.5-VL-7B & \xmark & 84.3 & 13.4 & 84.1 & 29.6 & 83.7 & 23.1 & 84.0 & 24.1 \\
Qwen2.5-VL-7B & \cmark & 85.0 & 12.5 & 84.4 & 28.9 & 84.2 & 19.9 & 84.2 & 22.5 \\
Qwen3-VL-8B & \xmark & 85.4 & 14.5 & 84.6 & 32.0 & 84.0 & 18.3 & 84.5 & 23.7 \\
Qwen3-VL-8B & \cmark & 85.8 & 16.4 & 84.6 & 28.0 & 84.3 & \textbf{23.3} & 84.7 & 24.1 \\
\midrule
Qwen3-VL-8B-FT & \xmark & 87.2 & 19.1 & 87.2 & 34.2 & 87.0 & 20.5 & 87.1 & 26.4 \\
\textbf{Qwen3-VL-8B-FT} & \cmark & \textbf{87.2} & \textbf{24.3} & \textbf{87.4} & \textbf{37.0} & \textbf{87.2} & 21.5 & \textbf{87.2} & \textbf{29.1} \\
\bottomrule
\end{tabular}
\caption{Category-conditioned generation with BERTScore and Content Consistency (\%); EIC-Gen denotes event conditioning.}
\label{tab:main_quality}
\end{table*}

\section{Method}

\subsection{Event-Informed Commentary Generation}
Event-Informed Commentary Generation (EIC-Gen) converts structured punch events into natural-language cues for Category-Conditioned Generation. It verbalizes each event in a fixed order: timestamp and boxer, followed by available technique, effectiveness and target. Effectiveness selects the main verb; for example, ``[10.1s] Red, left hook, landed on the torso.'' We time-order the $k_e=16$ latest events, concatenate them with recent commentary history, and feed them into the MLLM. Historical replay selects cached events by start time. The appendix gives all attribute-to-phrase templates.

\begin{figure}[!t]
    \centering
    \includegraphics[width=0.88\linewidth]{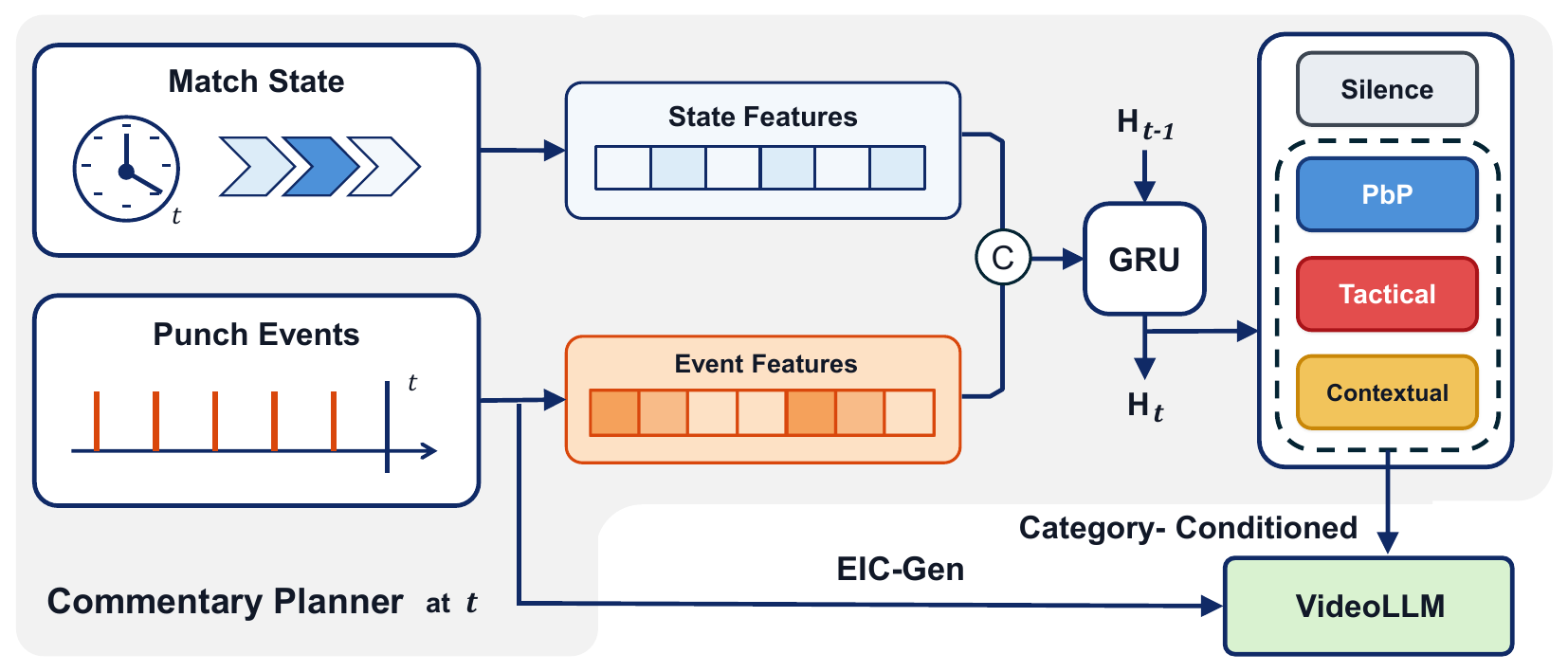}
    \caption{Commentary Planner. Non-Silence predictions condition the fine-tuned EIC-Gen generator, enabling autonomous streaming commentary.}
    \label{fig:planner}
\end{figure}

\subsection{Commentary Planner}
EIC-Gen produces category-conditioned content, while the Commentary Planner decides whether to speak and which category to use (Figure~\ref{fig:planner}). At each second $t$, a 10-dimensional Match State encodes video and match-unit progress, phase, and time remaining from the manifest. The Commentary Planner also aggregates punch counts, effectiveness confidence, and event recency into a 15-dimensional summary using event records whose end time precedes $t$; the appendix lists the complete features and objectives.

The event summary is mapped to 16 dimensions and concatenated with Match State before a single-layer GRU with 64 hidden units. Its joint head predicts four states $\{\text{Silence},\text{PbP},\text{Tactical},\text{Contextual}\}$. At non-silence steps, the predicted category conditions the sentence generator.

\section{Experiments}

\subsection{Experimental Setup}
\textbf{Content Baselines.} We evaluate GPT-4o-mini, LLaVA-OV1.5-7B, InternVL3.5-8B, Qwen2.5-VL-7B and Qwen3-VL-8B \cite{an2025llava,wang2025internvl3,bai2025qwen3}.

\textbf{Streaming Baselines.} We evaluate released StreamingVLM, LiveCC, and Proact-VL in their native configurations under a unified output and scoring contract. Random and Uniform sample commentary timestamps randomly or uniformly according to BoxComm-Train commentary rates and use the same fine-tuned generator as our Planner; Planner w/o Event directly ablates past event input.

\textbf{Implementation Details.} We fine-tune Qwen3-VL-8B-Instruct on BoxComm-Train for sentence generation using rank-64 LoRA for 2 epochs; the appendix gives the full configuration. The Planner trains on whole-video one-second sequences and resets per video. GT onsets label speaking, unoccupied bins label Silence, and occupied bins are masked. Training combines class-balanced focal loss with soft temporal-IoU and event-consistency objectives.

\textbf{Metrics.} For category-conditioned generation, we report BERTScore (F1) and GPT-based Content Consistency Accuracy. For commentary rhythm, we jointly report Temporal-IoU (t-IoU) and silence-aware four-state JS divergence.

\subsection{Human Validation}
\label{sec:human_validation}
We validate the automatic annotations and GPT-based evaluator through a 30-round human audit. From 40 evaluation matches, we sample ten rounds from each of Rounds 1, 2, and 3, covering 1,198 commentary sentences and 5,141 punch candidates. Two annotators independently recheck 10 overlapping rounds. Table~\ref{tab:human_validation} summarizes the results; the appendix provides the complete protocol and counts.

\begin{table}[!t]
\centering
\footnotesize
\renewcommand{\arraystretch}{1.12}
\begin{tabular*}{\linewidth}{@{\extracolsep{\fill}}lcc@{}}
\toprule
\textbf{Measure} & \shortstack{\textbf{30-round}\\\textbf{human check}} & \shortstack{\textbf{10-round}\\\textbf{A/B recheck}} \\
\midrule
Category & 90.23 Acc. & 93.46 \\
GPT Consistency & 92.82 Acc. & 82.94 \\
Punch Detection & 97.06 F1 (94.57 / 99.67) & 92.92 \\
Punch Effect & 97.44 F1 (99.67 / 95.30) & 96.51 \\
Punch Attributes & 99.20 Acc. & 99.03 \\
\bottomrule
\end{tabular*}
\caption{Human validation and A/B recheck (\%).}
\label{tab:human_validation}
\end{table}

\textbf{Commentary labels and judge.} GPT-5.2 matches human-corrected categories on 90.23\% of sentences. Of the 117 errors, 86 involve confusion between Play-by-Play and Tactical, so most ambiguity lies at the boundary between these categories. Content Consistency agrees with human judgments on 92.82\% of pairs. Most remaining errors are lenient over-acceptances; the reference-independent human evaluation provides a complementary check, with full counts in the appendix.

\textbf{Punch pseudo-labels.} Detection and effective-punch classification reach 97.06\% and 97.44\% F1, respectively, while the four attributes average 99.20\% accuracy. These results support the use of extracted events as fine-grained cues for commentary grounding and timing analysis.

On the 10 overlaps, Category and Consistency agreement reach 93.46\% ($\kappa=.890$) and 82.94\% ($\kappa=.598$); 20 of 28 Category disagreements reflect ontology boundaries. Together, the audits support benchmark-scale evaluation; most remaining uncertainty comes from commentary interpretation rather than event extraction.

\subsection{Quantitative Results}
Table~\ref{tab:main_quality} shows model-dependent EIC-Gen gains. It improves GPT-4o-mini and LLaVA-OV1.5-7B by 3.1 and 10.4 Overall points, with a smaller gain for Qwen3-VL (+0.4) and declines for InternVL3.5 and Qwen2.5-VL. With Qwen3-VL, fine-tuning raises Overall from 24.1\% to 29.1\% (+7.9 PbP; +9.0 Tactical).

Within fine-tuned Qwen3-VL, events raise Overall from 26.4\% to 29.1\% (+5.2 PbP; +2.8 Tactical; +1.0 Contextual), while BERTScore stays nearly flat. Semantic similarity alone therefore cannot verify the same boxing facts. Contextual accuracy is 21.5\% after fine-tuning and 23.3\% before. In a separate audit, 35 of 100 Contextual references lacked support in the supplied video and history (appendix).

Table~\ref{tab:input_ablation} shows that removing video costs 3.7 PbP points but changes Tactical and Contextual modestly, suggesting that video is used mainly for immediate descriptions. Removing history cuts Tactical/Contextual from 37.0/21.5\% to 14.3/9.3\%, showing their strong reliance on prior commentary. Event gains saturate at 16; the appendix reports the history-only control.

\begin{table}[!t]
\centering
\small
\renewcommand{\arraystretch}{1.10}
\begin{tabular*}{\linewidth}{@{\extracolsep{\fill}}lrrrr@{}}
\toprule
\textbf{Input} & \textbf{PbP} & \textbf{Tac} & \textbf{Ctx} & \textbf{Overall} \\
\midrule
Full (16 Events) & 24.3 & \textbf{37.0} & 21.5 & \textbf{29.1} \\
w/o Video & 20.6 & 35.5 & 20.1 & 27.2 \\
w/o History & 17.5 & 14.3 & 9.3 & 13.2 \\
w/o Events & 19.1 & 34.2 & 20.5 & 26.4 \\
\midrule
w/ 4 Events & 22.5 & 34.0 & 20.8 & 27.1 \\
w/ 8 Events & 23.6 & 34.7 & 21.5 & 27.9 \\
w/ 32 Events & \textbf{24.4} & 36.9 & \textbf{21.7} & \textbf{29.1} \\
\bottomrule
\end{tabular*}
\caption{Content Consistency input ablation (\%).}
\label{tab:input_ablation}
\end{table}
\FloatBarrier

\begin{table}[!t]
\centering
\small
\renewcommand{\arraystretch}{1.12}
\begin{tabular*}{\linewidth}{@{\extracolsep{\fill}}l|cccc|c@{}}
\toprule
\multirow{2}{*}{\textbf{Model}} & \multicolumn{4}{c|}{\textbf{t-IoU$\uparrow$}} & \multirow{2}{*}{\textbf{JS$\downarrow$}} \\
\cmidrule{2-5}
 & \textbf{PbP} & \textbf{Tac} & \textbf{Ctx} & \textbf{Mean} &  \\
\midrule
LiveCC & 0.043 & 0.199 & 0.130 & 0.124 & 0.315 \\
StreamingVLM & 0.047 & 0.152 & 0.105 & 0.101 & 0.255 \\
Proact-VL & 0.053 & 0.167 & \textbf{0.131} & 0.117 & 0.263 \\
\midrule
Random & 0.033 & 0.176 & 0.064 & 0.091 & 0.245 \\
Uniform & 0.057 & 0.196 & 0.074 & 0.109 & 0.274 \\
\textbf{Our Planner} & \textbf{0.105} & 0.209 & 0.111 & \textbf{0.141} & \textbf{0.187} \\
\midrule
Planner w/o Event & 0.086 & \textbf{0.212} & 0.103 & 0.134 & 0.188 \\
GT Ref.$^\dagger$ & 0.396 & 0.506 & 0.441 & 0.447 & 0.070 \\
\bottomrule
\end{tabular*}
\caption{Four-state commentary rhythm; GT Ref.$^\dagger$ uses GT timing, category, and history.}
\label{tab:rhythm}
\end{table}
\FloatBarrier

Table~\ref{tab:rhythm} shows that our Planner raises Mean t-IoU from 0.124 to 0.141 and reduces JS to 0.187. GT Ref. uses GT timestamps and categories, but generated lengths, interval reconstruction, and realized category labels yield 0.447 Mean t-IoU (appendix). Random and Uniform match training-set timestamp density without category conditions; they favor Tactical and miss PbP/Contextual rhythm, showing that frequency priors are insufficient. Removing events mainly lowers PbP from 0.105 to 0.086 while JS is nearly unchanged: Match State captures broad rhythm, whereas events refine short-timescale placement.

\begin{figure}[!t]
\centering
\includegraphics[width=\linewidth]{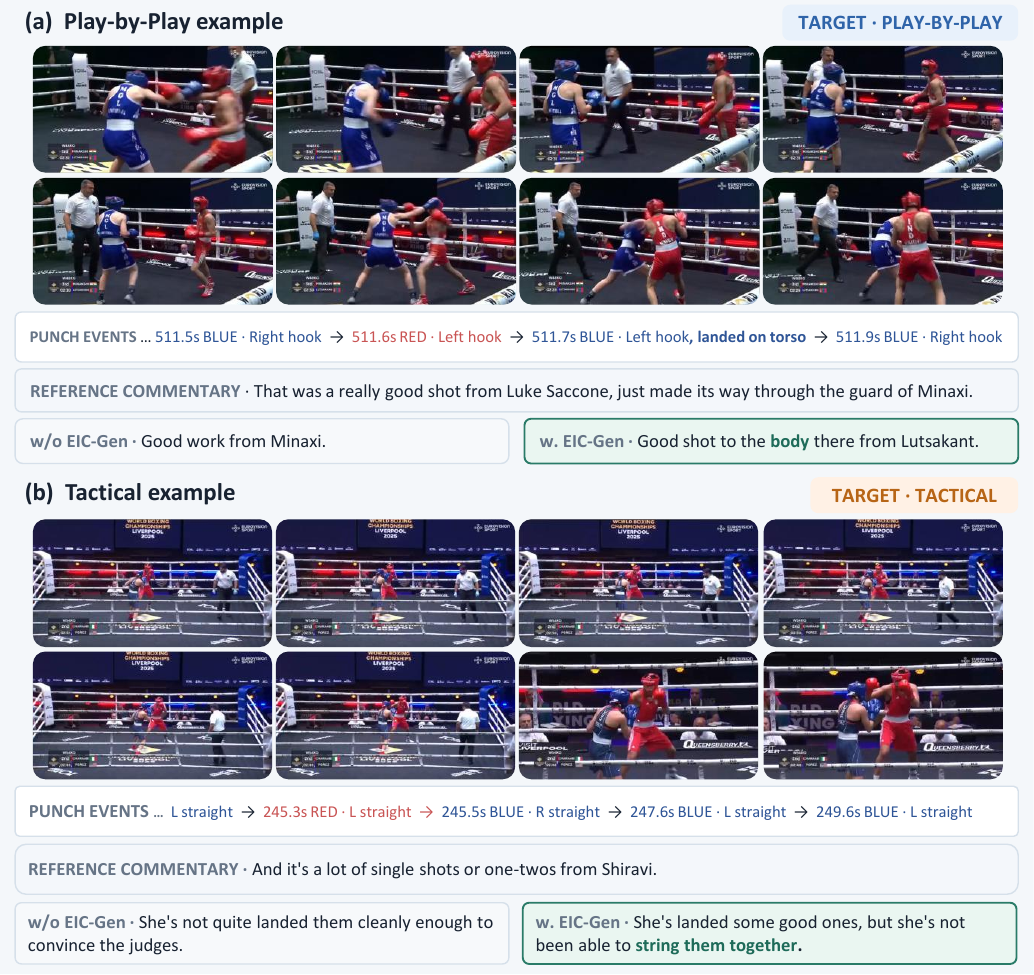}
\caption{Qualitative results without and with EIC-Gen.}
\label{fig:examples}
\small
\renewcommand{\arraystretch}{1.10}
\begin{tabular*}{\linewidth}{@{\extracolsep{\fill}}lrrrr@{}}
\toprule
\textbf{Model} & \textbf{Ground.} & \textbf{Contin.} & \textbf{Timing (\%)} & \textbf{Overall (\%)} \\
\midrule
LiveCC & 1.74 & 1.60 & 18.7 & 21.8 \\
Proact-VL & 1.81 & 1.62 & 21.7 & 22.8 \\
\textbf{Ours} & \textbf{2.12} & \textbf{1.72} & \textbf{59.6} & \textbf{55.4} \\
\bottomrule
\end{tabular*}
\captionof{table}{Human ratings on 90 windows. Preferences exclude Tie and None.}
\label{tab:e2e_human}
\end{figure}

\subsection{End-to-End Human Evaluation}
Three independent raters score 90 anonymized 20-second windows (270 judgments) with at least two onsets per system, natural timing, and randomized order (Table~\ref{tab:e2e_human}). They rate Groundedness (video/context support) and Continuity (coherence without repetition) from 1--3, then select the preferred Timing and Overall outputs. The evaluation does not require reference matching.

Ours is more grounded (2.12 versus 1.81). Among single-system selections, it receives 121/203 timing votes (59.6\%) and 112/202 overall votes (55.4\%). The result indicates that the system selects better-supported content and places it at more appropriate moments. Round-clustered bootstrap confirms all three advantages over both baselines (appendix). The 1.72/3 Continuity score leaves room to reduce formulaic phrasing and improve longer-range flow.

\subsection{Qualitative Analysis}
In Figure~\ref{fig:examples}(a), the event-conditioned output describes a body shot. In the Tactical example, the event sequence supports an exchange-level interpretation of isolated scoring punches.

\section{Conclusion}
BoxComm introduces the first category-aware boxing commentary benchmark, with 445 broadcasts, 52K aligned sentences, and 260K punch events. Its protocols separate content generation from speaking rhythm. EIC-Gen improves content consistency, while the Planner improves streaming rhythm. Human raters prefer the combined system for timing and overall quality. The two-stage design also keeps content and timing separately controllable. Our analysis shows that history supports Tactical and Contextual commentary, whereas recent events mainly benefit Play-by-Play and Tactical. Future work will improve knowledge grounding, long-range continuity, and cross-sport coverage.

% \begin{table}
%   \caption{Frequency of Special Characters}
%   \label{tab:freq}
%   \begin{tabular}{ccl}
%     \toprule
%     Non-English or Math&Frequency&Comments\\
%     \midrule
%     \O & 1 in 1,000& For Swedish names\\
%     $\pi$ & 1 in 5& Common in math\\
%     \$ & 4 in 5 & Used in business\\
%     $\Psi^2_1$ & 1 in 40,000& Unexplained usage\\
%   \bottomrule
% \end{tabular}
% \end{table}

% \begin{table*}
%   \caption{Some Typical Commands}
%   \label{tab:commands}
%   \begin{tabular}{ccl}
%     \toprule
%     Command &A Number & Comments\\
%     \midrule
%     \texttt{{\char'134}author} & 100& Author \\
%     \texttt{{\char'134}table}& 300 & For tables\\
%     \texttt{{\char'134}table*}& 400& For wider tables\\
%     \bottomrule
%   \end{tabular}
% \end{table*}

% Identification of funding sources and other support, and thanks to
% individuals and groups that assisted in the research and the
% preparation of the work should be included in an acknowledgment
% section, which is placed just before the reference section in your
% document.

% This section has a special environment:
% \begin{verbatim}
%   \begin{acks}
%   ...
%   \end{acks}
% \end{verbatim}
% so that the information contained therein can be more easily collected
% during the article metadata extraction phase, and to ensure
% consistency in the spelling of the section heading.

% Authors should not prepare this section as a numbered or unnumbered {\verb|\section|}; please use the ``{\verb|acks|}'' environment.


\clearpage
\section{Acknowledgments}
This research was supported by Beijing Natural Science Foundation (L252171) and Huawei’s AI Hundred Schools Program and was carried out using the Huawei Ascend AI technology stack. Additionally, we would like to acknowledge the Xinjiang Uygur Autonomous Region Sports Science Research Center and the research group led by Prof. Qingmin Fan at Beijing Sport University for their critical assistance with the data collection and annotation iteration processes.

\bibliography{references}

\clearpage
\appendix
\setcounter{figure}{0}
\setcounter{table}{0}
\setcounter{equation}{0}
\renewcommand{\thefigure}{A\arabic{figure}}
\renewcommand{\thetable}{A\arabic{table}}
\renewcommand{\theequation}{A\arabic{equation}}
\input{appendix}

\end{document}

%% file: appendix.tex
\section*{Technical Appendix}
This appendix provides extended generation and streaming analyses, implementation details for EIC-Gen and the Commentary Planner, annotation and human-evaluation protocols, the release scope, and the complete model-call prompts. The streaming sections distinguish the native end-to-end results in the main paper from controlled plan-only, length-controlled, and common-realizer analyses.

\section{Extended Generation Experiments}

\subsection{History, Video, and Event Inputs}
Table~\ref{tab:supp_history_controls} reports matched input controls of the same fine-tuned generator on all 5,588 targets. Each arm receives the target category. All arms were scored in the same GPT-5.2 evaluation batch. History provides substantial discourse context. Adding video to history raises Overall Content Consistency from 25.0\% to 26.4\% and Play-by-Play from 17.1\% to 19.1\%.

\begin{table}[!b]
\centering
\footnotesize
\begin{tabular*}{\linewidth}{@{\extracolsep{\fill}}lrrrr@{}}
\toprule
\textbf{Inputs with target category} & \textbf{PbP} & \textbf{Tac} & \textbf{Ctx} & \textbf{All} \\
\midrule
Category only & 3.9 & 18.5 & 13.5 & 13.9 \\
History & 17.1 & 33.9 & 18.0 & 25.0 \\
Video & 8.3 & 9.1 & 8.8 & 8.8 \\
Events & 7.2 & 29.6 & 6.8 & 17.2 \\
Video and history & 19.1 & 34.5 & 20.0 & 26.4 \\
\bottomrule
\end{tabular*}
\caption{Content Consistency (\%) in the matched history and perception diagnostic. The denominators are 1,090 PbP, 2,524 Tactical, and 1,974 Contextual targets.}
\label{tab:supp_history_controls}
\end{table}

\subsection{History Context Beyond Direct Copying}

The main-paper input ablation reports the canonical Content Consistency results for video, commentary history, and past-event inputs. Here we examine why preceding commentary is useful. For a target sentence with onset $t$, reference history contains the eight preceding sentences whose end timestamps do not exceed $t$, in chronological order. The target and all later commentary are excluded. The target category is supplied in every condition. This controlled sentence-level protocol uses reference history, whereas autonomous streaming uses each system's own generated history.

We first examine whether the history advantage can be explained by direct repetition. A target is marked as a near duplicate if any history sentence is an exact normalized match or reaches 0.80 token-set Jaccard, 0.90 token F1, or 0.90 ROUGE-L F1. This occurs for only 14 of 5,588 targets (0.25\%; 13 exact matches). Table~\ref{tab:supp_history_overlap} reports the paired token-F1 difference after removing these targets from both conditions.

The gain remains positive for every category after near-duplicate removal. A complementary semantic check uses BERTScore v0.3.12 with \texttt{roberta-large} (layer 17; no IDF or baseline rescaling). Only 26/5,588 targets (0.47\%) exceed the 0.95 near-duplicate threshold. Removing them gives .8574 for Full and .8510 for w/o History, a paired difference of $+.0064$ [$+.0051,+.0077$]; the category differences are $+.0090$, $+.0086$, and $+.0022$. These lexical and semantic checks show that history contributes discourse context beyond direct repetition of the target.

\begin{center}
\centering
\footnotesize
\setlength{\tabcolsep}{4.5pt}
\begin{tabular*}{\columnwidth}{@{\extracolsep{\fill}}lrr}
\toprule
Category & Near duplicates & Full $-$ w/o History [95\% CI] \\
\midrule
PbP        & 2/1,090 (0.18\%)  & $+.0380$ [$+.0205,+.0556$] \\
Tactical   & 10/2,524 (0.40\%) & $+.0270$ [$+.0204,+.0335$] \\
Contextual & 2/1,974 (0.10\%)  & $+.0181$ [$+.0098,+.0268$] \\
Overall    & 14/5,588 (0.25\%) & $+.0260$ [$+.0208,+.0311$] \\
\bottomrule
\end{tabular*}
\captionof{table}{History overlap and clean-set token-F1 differences. Confidence intervals use match-clustered bootstrap resampling.}
\label{tab:supp_history_overlap}
\end{center}

\subsection{Additional EIC-Gen Examples}

\begin{figure*}[t]
\centering
\includegraphics[width=\textwidth]{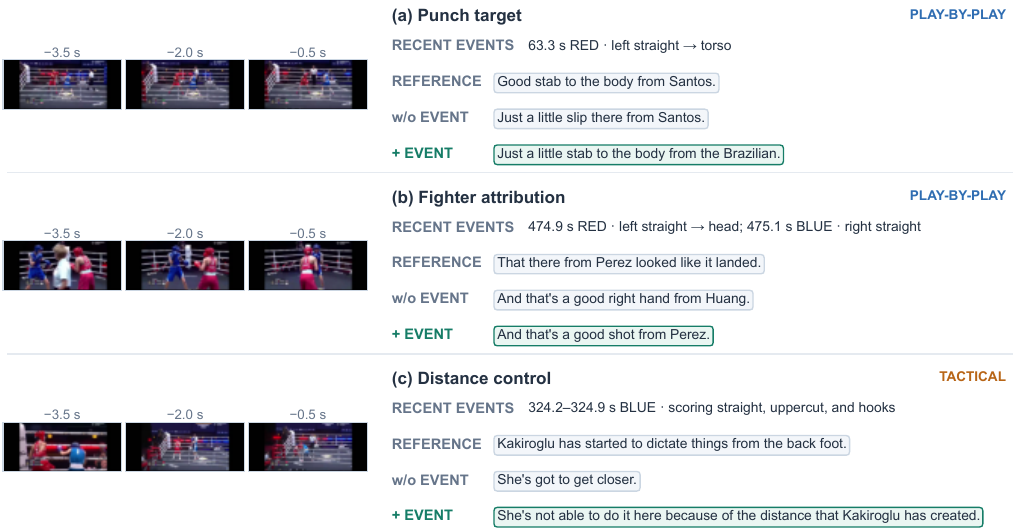}
\caption{Additional paired EIC-Gen examples. Event cues improve punch target and fighter attribution in Play-by-Play, and connect a scoring sequence to Tactical distance control.}
\label{fig:supp_eicgen_examples}
\end{figure*}

Figure~\ref{fig:supp_eicgen_examples} complements the aggregate results with three paired checkpoint-2888 examples. The video, target category, reference history, and generator are identical within each pair; only the recent past-event cue changes. The examples illustrate how structured event cues can affect generated details and tactical interpretation.

\section{Streaming Evaluation Details}
\label{sec:supp_streaming}

\paragraph{Realized speech intervals.}
Each autonomous method retains its own sentence onset. For a final text containing $N$ words, audible duration follows a common 2.5-words-per-second convention with a one-second minimum. An interval ends no later than the next sentence onset or the video boundary, preventing overlapping speech. When an interval is truncated, only the text prefix that can be realized within the remaining duration is retained and passed to the shared GPT-5.2 category classifier. Method-specific onset reconstruction is summarized below.

Reference intervals retain the word-level alignment produced during transcript processing: each final sentence starts at the first word in its aligned span and ends at the last. When a long segment is split at an internal word boundary, the two resulting sentences inherit the word times on their respective sides of that boundary.

\paragraph{t-IoU aggregation.}
For each match $m$ and category $c$, the symmetric max-overlap score in the main paper is computed only between intervals from that match. The final category score first averages the 40 match-level values equally, and the reported Mean then averages the three categories. If either predicted or reference intervals are empty for a match--category pair, its value is defined as zero. Table~\ref{tab:supp_length_controlled} complements this interval-coverage metric by controlling predicted duration.

For four-state JS, each video is divided into consecutive 60-second cells, retaining the final partial cell; this yields 436 cells with positive duration. Overlapping intervals are resolved by the latest onset, and uncovered time is Silence. For each cell, we compute the base-2 Jensen--Shannon divergence in Eq.~(2) of the main paper between reference and predicted duration distributions over \{Silence, Play-by-Play, Tactical, Contextual\}, then average cells equally. The frozen main-paper JS reference uses 5,563 serialized human-text intervals and their GPT-5.2 labels. Main-paper t-IoU instead uses the original 5,588 timestamped reference intervals and their annotation labels.

\begin{center}
\footnotesize
\begin{tabular*}{\linewidth}{@{\extracolsep{\fill}}l l p{0.46\linewidth}@{}}
\toprule
\textbf{Method} & \textbf{Adapt.} & \textbf{Timing and system} \\
\midrule
Released systems & None & Native outputs and released backbones \\
Random / Uniform & Gen. & TRAIN-rate schedules; shared EIC-Gen \\
Our Planner & Both & Planner onsets; shared EIC-Gen \\
w/o Event & Both & Direct Planner input ablation \\
GT Ref. & Gen. & GT onset/category/history; shared EIC-Gen \\
\bottomrule
\end{tabular*}
\captionof{table}{Streaming comparison scope. ``Adapt.'' describes the evaluated checkpoint.}
\label{tab:supp_streaming_setup}
\end{center}

``Released systems'' denotes Proact-VL, LiveCC, and StreamingVLM evaluated with their native streaming objectives and interfaces. Random, Uniform, Our Planner, and Planner w/o Event share the fine-tuned EIC-Gen generator; Random and Uniform derive timestamp schedules from BoxComm-Train commentary rates. The primary comparison applies common scoring metrics and classifier definitions. The original GT Ref.\ row uses its frozen reconstructed intervals and full-sentence labels, as detailed below. The length-controlled diagnostic uses the later unified audible-prefix conversion for GT Ref.

LiveCC uses \texttt{LiveCC-7B-Instruct} at one-second updates with two frames per second, persistent native history, and at most 20 new tokens per update. StreamingVLM also updates once per second, using a four-second visual window and its released persistent-KV configuration. Proact-VL uses \texttt{proactvl\_base\_liveccbase} at one-second updates, its released state threshold of 0.5, and at most 12 new tokens. These systems are evaluated with their released native streaming configurations.

\paragraph{Method-specific onset reconstruction.}
For LiveCC, native one-second streaming chunks are joined until terminal punctuation or video end; the sentence starts at the first non-silent chunk contributing text. For StreamingVLM, nonempty entries in \texttt{responses} are joined in the same way, using the first contributing \texttt{start\_time}; literal ellipses are treated as silence. For Proact-VL, consecutive active fragments are joined using its released streaming boundaries, and the first active fragment supplies the native onset. Uniform repeats the BoxComm-Train mean inter-commentary interval from a half-interval offset; Random samples the corresponding per-video number of onsets uniformly with a fixed seed and sorts them. For Ours and w/o Event, an accepted non-Silence decision on the one-second Planner grid is the sentence onset. GT Ref. uses the reference segment start time to trigger the shared generator with privileged category and history. The autonomous rows use the shared audible-prefix conversion. The GT Ref.\ row in Table~4 of the main paper retains its original reconstructed intervals and full-sentence labels.

Autonomous methods use their own history or native state. GT Ref. uses professional reference history as an additional controlled input.

\paragraph{Privileged generator reference.}
GT Ref. supplies the generator with ground-truth onset, target category, and previous-commentary history, while still generating a new sentence. Differences in sentence length and realized category keep its t-IoU below 1. The ordered decomposition in Table~\ref{tab:supp_gt_decomp} starts with exact interval and category identity on all 5,588 reference utterances. Among these same-onset pairs, 38.4\% of generated and reference durations differ by more than a factor of two. Generated lengths, sentence reconstruction, and realized categories each change overlap. Reconstruction merges 172 raw slots into adjacent slots and leaves 5,416 sentences. The final category step combines generator behavior and the classifier's decisions.

\begin{table}[!t]
\centering
\footnotesize
\begin{tabular*}{\linewidth}{@{\extracolsep{\fill}}lrr@{}}
\toprule
\textbf{Cumulative scoring change} & \textbf{Intervals} & \textbf{Mean t-IoU} \\
\midrule
Exact reference identity & 5,588 & 1.0000 \\
Generated endpoints & 5,588 & .5889 \\
Reconstructed support and endpoints & 5,416 & .5768 \\
Realized sentence labels & 5,416 & .4475 \\
\bottomrule
\end{tabular*}
\caption{Ordered decomposition of the original GT Ref.\ row on all 40 matches. GT onsets and target labels are held until the corresponding reconstruction or labeling step.}
\label{tab:supp_gt_decomp}
\end{table}

\subsection{Length-Controlled Timing Diagnostics}

To separate trigger timing from generated sentence length, fixed-duration t-IoU replaces every predicted end time with its onset plus the BoxComm-Train mean duration for that realized category (2.899, 3.353, and 3.160 seconds for Play-by-Play, Tactical, and Contextual), capped by the next onset and video end.

\begin{table}[ht]
\centering
\footnotesize
\begin{tabular*}{\linewidth}{@{\extracolsep{\fill}}lrrrr@{}}
\toprule
\textbf{Method} & \textbf{PbP} & \textbf{Tac} & \textbf{Ctx} & \textbf{Mean} \\
\midrule
Proact-VL & .0682 & .1799 & .1196 & .1226 \\
LiveCC & .0471 & .2108 & .1400 & .1327 \\
StreamingVLM & .0493 & .1499 & .1085 & .1026 \\
\midrule
Random & .0390 & .1847 & .0739 & .0992 \\
Uniform & .0633 & .2090 & .0928 & .1217 \\
\midrule
\textbf{Ours} & \textbf{.1261} & .2153 & \textbf{.1422} & \textbf{.1612} \\
w/o Event & .1009 & \textbf{.2206} & .1261 & .1492 \\
\midrule
GT Ref. & .4438 & .5729 & .5454 & .5207 \\
\bottomrule
\end{tabular*}
\caption{Length-controlled t-IoU after replacing generated duration with a TRAIN-derived category mean.}
\label{tab:supp_length_controlled}
\end{table}

Among autonomous systems, Ours attains the highest mean t-IoU (.1612). Match-clustered bootstrap differences are $+.0285$ [$+.0168,+.0407$] over LiveCC and $+.0120$ [$+.0052,+.0188$] over w/o Event. The temporal-overlap advantage persists with standardized durations. GT Ref.'s .5207 uses retained onsets, realized labels, and TRAIN durations. In that scoring representation, the corresponding realized-duration score is .5030. Using reference onsets and labels on all 5,588 reference slots with TRAIN mean durations gives .7435, reflecting imperfect reconstruction of individual reference endpoints.

\subsection{Common-Generator Schedule Isolation}

We additionally replay the frozen onset and category sequences of LiveCC, StreamingVLM, Proact-VL, and our Planner through the same checkpoint-2888 EIC-Gen realizer. Every arm uses deterministic decoding, the latest 16 past events, previous-eight generated self-history, and the same audible serialization. Baseline categories are the frozen labels of their original outputs, while our Planner supplies its predicted category. The resulting comparison focuses on the schedule/category interface under a common realizer.

Ours achieves the highest Mean t-IoU and the lowest four-state JS. Its paired Mean gains over LiveCC, StreamingVLM, and Proact-VL are $+.0243$ [$+.0149,+.0345$], $+.0574$ [$+.0487,+.0659$], and $+.0349$ [$+.0229,+.0465$], respectively. LiveCC remains stronger on Tactical overlap, whereas Ours leads on Play-by-Play and Contextual. Output densities show distinct speaking patterns: LiveCC speaks more frequently and Proact-VL less frequently than Ours.

\begin{figure}[ht]
\centering
\includegraphics[width=\columnwidth]{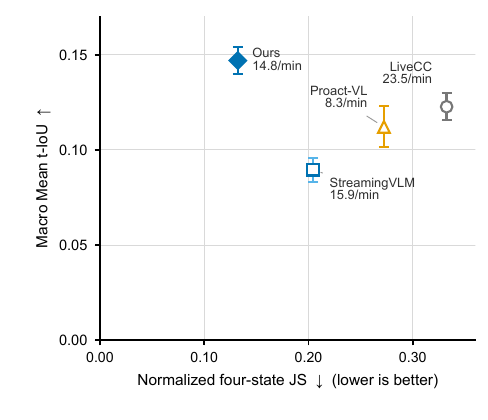}
\caption{Common-realizer schedule comparison on EVAL40. Higher Macro Mean t-IoU and lower normalized four-state JS are better; vertical bars are 40-match bootstrap 95\% confidence intervals, and labels report audible outputs per minute. All methods use the same checkpoint-2888 EIC-Gen realizer.}
\label{fig:supp_schedule_isolation}
\end{figure}

\begin{table}[ht]
\centering
\footnotesize
\begin{tabular*}{\columnwidth}{@{\extracolsep{\fill}}lrrrrr@{}}
\toprule
\textbf{Schedule} & \textbf{PbP} & \textbf{Tac} & \textbf{Ctx} & \textbf{Mean} & \textbf{JS}$\downarrow$ \\
\midrule
LiveCC & .0540 & \textbf{.2061} & .1082 & .1227 & .3325 \\
StreamingVLM & .0352 & .1398 & .0939 & .0896 & .2044 \\
Proact-VL & .0619 & .1738 & .1005 & .1121 & .2724 \\
\textbf{Ours} & \textbf{.1233} & .1839 & \textbf{.1338} & \textbf{.1470} & \textbf{.1323} \\
\bottomrule
\end{tabular*}
\caption{Schedule/category-interface comparison with a common EIC-Gen realizer on EVAL40. Paired confidence intervals and output densities are reported in the text and Figure~\ref{fig:supp_schedule_isolation}.}
\label{tab:supp_common_generator}
\end{table}

\FloatBarrier
\subsection{StreamingVLM Domain Adaptation}

We adapt the released StreamingVLM on 27,262 windows of up to eight seconds from 345 BoxComm training matches, retaining silent windows and native streaming supervision. The visual encoder is frozen and the language model uses LoRA (rank 64, $\alpha=128$, dropout .05). Training applies uniform cross-entropy over assistant body and EOS tokens with effective batch size 32, a peak learning rate of $10^{-5}$, 60 warm-up steps, and cosine decay. We report the fixed 2,000-step checkpoint from one run. Both variants use native decoding without category prompts, punch events, or Match State. We assign output categories after generation using the same classifier. Table~\ref{tab:supp_adapt_rhythm} compares both variants under identical full-match inference and rhythm scoring. The released result in this paired run differs from the historical result in the main comparison. Paired match-bootstrap 95\% intervals for adapted minus released are $[-.0190,.0002]$ for Mean t-IoU and $[.0242,.0753]$ for JS.

\begin{table}[!t]
\centering
\footnotesize
\begin{tabular*}{\linewidth}{@{\extracolsep{\fill}}lrrrrr@{}}
\toprule
\textbf{Variant} & \textbf{PbP} & \textbf{Tac} & \textbf{Ctx} & \textbf{Mean} & \textbf{JS} \\
\midrule
Released & .0596 & .1611 & .1052 & .1086 & .2122 \\
Adapted & .0475 & .1487 & .1012 & .0991 & .2619 \\
\bottomrule
\end{tabular*}
\caption{Matched StreamingVLM rhythm comparison across all 40 evaluation matches. Adapted denotes the 2,000-step checkpoint. The four t-IoU columns are followed by four-state JS.}
\label{tab:supp_adapt_rhythm}
\end{table}

For content, we match realized and reference intervals one-to-one using positive temporal overlap, then retain same-category targets that pass the boundary filter. Released and adapted variants cover 1,123 and 650 same-category targets, respectively. Their intersection contains 219 targets (3 PbP, 168 Tactical, 48 Contextual). Under the same GPT-5.2 rubric, the released and adapted systems score 13/219 (5.94\%) and 28/219 (12.79\%). On these reference IDs, category-conditioned Qwen3-VL and full EIC-Gen score 53/219 (24.20\%) and 66/219 (30.14\%). The two anchors receive a category prompt and professional history, while StreamingVLM outputs follow autonomous timing and self-history. This subset shows a content gain on jointly covered targets alongside weaker aggregate rhythm after adaptation.

\section{Model and Training Details}

The evaluated models use video without broadcast audio or ASR features. The live extractor classifies attributes from completed segments and emits each record after classification. The historical Planner used cached event end times as observation timestamps. Actual online use requires gating events on arrival, since classification can finish after the recorded event end.

\paragraph{Event verbalization.}
A deterministic template converts each selected event into a compact natural-language cue. The frozen historical rhythm generator selects cached events with start time earlier than the trigger. It can therefore include records that would be unfinished at that time. The live replay uses emitted records. A release-time-gated rerun would be needed to quantify the effect on the historical end-to-end scores. The cue records video time, fighter side, canonical punch technique, and supported landing information. Side is normalized to red or blue, technique combines left/right with straight, hook, or uppercut when available, and recognized head/chest/abdomen targets are normalized to head or torso. Unsupported or missing attributes are omitted rather than guessed. The most recent 16 past events are ordered chronologically and joined using fixed separators; when none is available, the event-context line is omitted. This structured conversion provides concise grounding cues while leaving the generator free to choose the final commentary wording.

For sentence-level generation, we adapt Qwen3-VL-8B-Instruct on BoxComm-Train using LoRA with rank 64, $\alpha=128$, and dropout 0.05 on the attention \texttt{q/k/v/o} projections and MLP \texttt{gate/up/down} projections. Optimization uses AdamW, a cosine learning-rate schedule with peak learning rate $10^{-4}$, batch size 2, and gradient accumulation 8. We use the step-2888 checkpoint at epoch 2. At evaluation, each four-second query clip is represented by eight uniformly sampled frames, and decoding is deterministic (temperature 0, top-$p$ 1, and at most 64 new tokens). A same-protocol reproduction required approximately 12.35 hours on two A800 GPUs.

\paragraph{Match State.}
Match phase and interval boundaries are read from the evaluation manifest. Let $T$ be the video duration and $[a,b)$ the current round, intermission, or post-match segment. With $u=t/\max(T,1)$ and $v=(t-a)/\max(b-a,1)$, the 10-dimensional state is
\begin{equation}
\begin{aligned}
s_t=[&u,\sin 2\pi u,\cos 2\pi u,
v,\sin 2\pi v,\cos 2\pi v,\\
&I_{\mathrm{round}},I_{\mathrm{inter}},
I_{\mathrm{post}},1-v].
\end{aligned}
\end{equation}
Continuous features are clipped to their natural ranges and standardized with BoxComm-Train statistics. Each phase indicator is 1 when its phase is active and 0 otherwise. The Event-only diagnostic in Table~\ref{tab:supp_planner_inputs} removes all Match State fields while retaining past-event inputs.

\paragraph{Fixed-schedule Match State.}
On the same 40 matches, we replace all ten fields with elapsed-time features from three consecutive 180\,s rounds separated by two 60\,s breaks, starting at the original clock origin. For $t\geq660$\,s, phase is post-match and $v=1$. The checkpoint, event inputs, decision rules, and evaluation horizons are held fixed. The recovered TRAIN scaler reproduces all 6,449 original onset--category tuples. Plan-only Mean t-IoU rounds to .151 for both schedules (.15081 and .15073; paired 95\% match-bootstrap interval for the difference $[-.00169,.00148]$), plan JS from .12019 to .11885, and per-second state agreement is 97.25\%. PbP overlap decreases from .12497 to .12209. The test isolates scheduled Match State and leaves event eligibility and generation outside the intervention.

\paragraph{Past-event summary.}
For each $h\in\{2,4,8,16\}$ seconds, events in $[t-h,t)$ yield three features:
\begin{equation}
\begin{aligned}
a_h&=\log(1+N_h),\\
b_h&=\log\!\left(1+\sum_i \operatorname{clip}(c_i,0,1)\right),\\
d_h&=\log(1+N_h^{\rm elig}),
\end{aligned}
\end{equation}
where $N_h$ counts punches, $c_i$ is effectiveness confidence, and $N_h^{\rm elig}$ counts events passing the upstream in-round validity flag. These 12 features are augmented by log-recencies of the latest event, eligible event, and predicted-effective event. Recencies are capped at 16 seconds (and set to 16 when absent); all 15 features use TRAIN-only standardization. The historical in-round flag is manifest-derived. Replaying phase-start notifications through event completion reproduced all 23,384 evaluation flags. In a live deployment the same flag can be derived from the latest observed phase transition.

\paragraph{Network and objectives.}
The event vector is projected by $\hat e_t=\operatorname{SiLU}(W_e e_t+b_e)\in\mathbb{R}^{16}$ and concatenated with $s_t$. A one-layer, 64-hidden-unit GRU updates $h_t$ and a linear joint head predicts Silence, Play-by-Play, Tactical, or Contextual. The Planner trains on whole-video one-second sequences and resets between videos. GT onsets label speaking, valid unoccupied bins label Silence, and bins occupied by ongoing speech are masked. Its objective combines class-balanced classification, differentiable category-wise temporal IoU, event-local consistency, and outside-event stability:
\begin{equation}
L=L_{\rm cls}+0.5L_{\rm tIoU}+0.25L_{\rm event}+0.05L_{\rm stable}.
\end{equation}
The classification term averages Event and event-clamped branches; each branch averages cross entropy and focal loss with $\gamma=1.5$. Square-root inverse-frequency weights are mean-normalized and multiplied by $[1,2.5,1,1]$ for Silence/PbP/Tactical/Contextual. $L_{\rm tIoU}$ uses class weights $[2.5,1,1]$. Within two seconds of a past event for PbP and eight seconds for Tactical, $L_{\rm event}$ trains the Event-minus-clamped probability residual toward the target residual. Outside the eight-second event neighborhood, $L_{\rm stable}$ averages symmetric KL divergence and the squared difference in non-Silence probability.

\Needspace{7\baselineskip}
\subsection{Pre-Realization Planner Analysis}

\paragraph{Inference.}
At each one-second step, a fixed keyed random stream samples Silence versus speaking with logit bias 2.4, then samples the commentary category from temperature-0.45 logits; Event features are scaled by 1.5. After a trigger of category $c$, the next trigger is delayed by $\max(3,1.4d_c)$ seconds, where $d_c=[2.394,2.783,2.482]$ for PbP, Tactical, and Contextual. The accepted category conditions EIC-Gen before the realized-speech conversion described above.

\begin{center}
\scriptsize
\begin{tabular*}{\linewidth}{@{\extracolsep{\fill}}lrrrrr@{}}
\toprule
\textbf{Input} & \textbf{PbP} & \textbf{Mean} & \textbf{JS}$\downarrow$ & \textbf{Onset} & \textbf{Fixed} \\
\midrule
\textbf{State + Event} & \textbf{.1250} & \textbf{.1508} & .1202 & \textbf{.1674} & \textbf{.1647} \\
State only & .1132 & .1418 & .1192 & .1541 & .1557 \\
Event only & .1161 & .1428 & \textbf{.1127} & .1585 & .1563 \\
\bottomrule
\end{tabular*}
\captionof{table}{Pre-realization Planner input attribution under the main seed. PbP and Mean are plan-level t-IoU before generator duration is realized; Onset is category-correct Macro-F1 at $\pm1$ second; Fixed uses TRAIN-derived category durations.}
\label{tab:supp_planner_inputs}
\end{center}

\paragraph{Seed stability.}
Repeating the full Planner with three fixed seeds gives pre-realization Plan Mean $.150871\pm.000125$, four-state JS $.120036\pm.000152$, onset Macro-F1 $.167203\pm.000186$, and fixed-duration Mean $.164581\pm.000168$ (mean $\pm$ sample standard deviation). Output density is identical across seeds at 14.7586 utterances per minute. The Planner is therefore stable across the tested random seeds.

\paragraph{Input attribution.}
Table~\ref{tab:supp_planner_inputs} separates Match State from past events. Event-only improves over State-only by .0011 Mean t-IoU, .0028 Play-by-Play t-IoU, and .0045 onset Macro-F1. Combining both sources improves over Event-only by .0080 Mean, .0089 Play-by-Play, and .0089 onset Macro-F1. Event-only obtains the lowest JS, indicating the closest aggregate four-state distribution, while the combined input gives substantially better temporal overlap and localization. Thus Match State and past events provide complementary scheduling evidence rather than interchangeable cues.

\FloatBarrier
\section{Annotation and Human Evaluation}

\subsection{Annotation Audit}

The audit covers 30 English evaluation rounds. Two annotators reviewed and corrected all automatic outputs within these intervals; 10 rounds were reviewed by both annotators. The Consistency audit therefore exhaustively covers all 1,198 commentary-level GPT-5.2 decisions (389 from Round 1 and 809 from Rounds 2--3), without item-level subsampling. These pairs come from the main fine-tuned EIC-Gen setting: Qwen3-VL-8B-FT checkpoint-2888 with the target category, video, previous-eight reference commentaries, and the latest 16 past events. Tables~\ref{tab:supp_commentary_audit} and~\ref{tab:supp_attributes} provide the corresponding confusion matrices and attribute results.

\begin{table}[ht]
\centering
\footnotesize
\textbf{(a) Commentary Category}\\[2pt]
\begin{tabular}{lrrr}
\toprule
\textbf{Human $\backslash$ GPT} & \textbf{PbP} & \textbf{Tac} & \textbf{Ctx} \\
\midrule
PbP & 191 & 49 & 9 \\
Tactical & 37 & 539 & 9 \\
Contextual & 2 & 11 & 351 \\
\bottomrule
\end{tabular}
\\[4pt]
\textbf{(b) Content Consistency}\\[2pt]
\begin{tabular}{lrr}
\toprule
\textbf{Human $\backslash$ GPT} & \textbf{Incons.} & \textbf{Cons.} \\
\midrule
Inconsistent & 809 & 84 \\
Consistent & 2 & 303 \\
\bottomrule
\end{tabular}
\caption{Commentary audit over 1,198 sentences. Rows are human-corrected references and columns are automatic outputs. For Consistency, ``Cons.'' is the positive class.}
\label{tab:supp_commentary_audit}
\end{table}

Of the 117 Category errors, 86 exchange Play-by-Play and Tactical. Punch Detection has TP/FP/FN $=4{,}862/279/16$; binary effective-punch classification has TP/FP/FN/TN $=893/3/44/3{,}922$.

For the punch audit, annotators reviewed 30 complete round clips alongside 5,141 automatically extracted candidates and could edit, delete, or add events, producing a 4,878-event corrected reference. Detection uses one-to-one matching by boxer side, punch-hand prefix, and temporal IoU $\geq0.5$. Effect is evaluated on the 4,862 pairs satisfying this matcher. Attribute accuracy instead follows 4,864 retained seed-to-human lineage pairs and counts missing final rows as errors; two retained events have corrected hand or temporal boundaries that break the detection matcher but preserve their event lineage.

Agreement between the two correction passes is 82.94\% on the 10 overlapping rounds, while GPT-versus-human accuracy is 92.82\% over all 1,198 audited pairs.

\paragraph{Primary-function labels.}
The three-way annotation records the primary communicative function of each segmented sentence. Semantic re-segmentation separates adjacent Play-by-Play description and Tactical interpretation when they form distinct utterances, while some mixed sentences remain close to the boundary. After guideline-based correction, the two annotators agree on 93.46\% of Category labels ($\kappa=.890$); 20 of the remaining 28 disagreements concern this boundary.

\begin{table}[ht]
\centering
\footnotesize
\begin{tabular*}{\linewidth}{@{\extracolsep{\fill}}lrr@{}}
\toprule
\textbf{Punch attribute} & \textbf{Accuracy} & \textbf{Agreement} \\
\midrule
Side & 99.47 & 100.00 \\
Distance & 99.12 & 98.77 \\
Technique & 98.58 & 97.61 \\
Target & 99.65 & 99.74 \\
\bottomrule
\end{tabular*}
\caption{Per-attribute audit results (\%). Accuracy uses all 30 rounds; agreement uses the 10 overlapping rounds.}
\label{tab:supp_attributes}
\end{table}

Punch Detection agreement compares the annotators' retain/remove decisions on the shared initial punch list. Effect and attribute agreement are computed on punches retained and aligned by both annotators.

\FloatBarrier
\subsection{1080p Replay Latency}
We ran two 30-second live replays at $1620\times1080$, with 20 seconds analyzed per replay, using one RTX 4090 for pose and event extraction and one for Qwen. Across all 10 accepted utterances, Planner-trigger-to-complete-text latency averaged 1.024\,s (range 0.810--1.174\,s), including queueing, frame wait, preprocessing, and decoding. In a separate event-stage measurement, complete event records reached the consumer an average of 186.5\,ms after the reference event end across 86 matched events out of 141 reference events, pooled across three replay windows. Trigger scheduling and deliberate silence are outside the generation measurement. If the Planner triggers as soon as an event is available, summing the two stage means gives an estimated 1.21\,s from event end to complete text.

\subsection{End-to-End Human Evaluation}

\begin{figure*}[t]
\centering
\includegraphics[width=\textwidth]{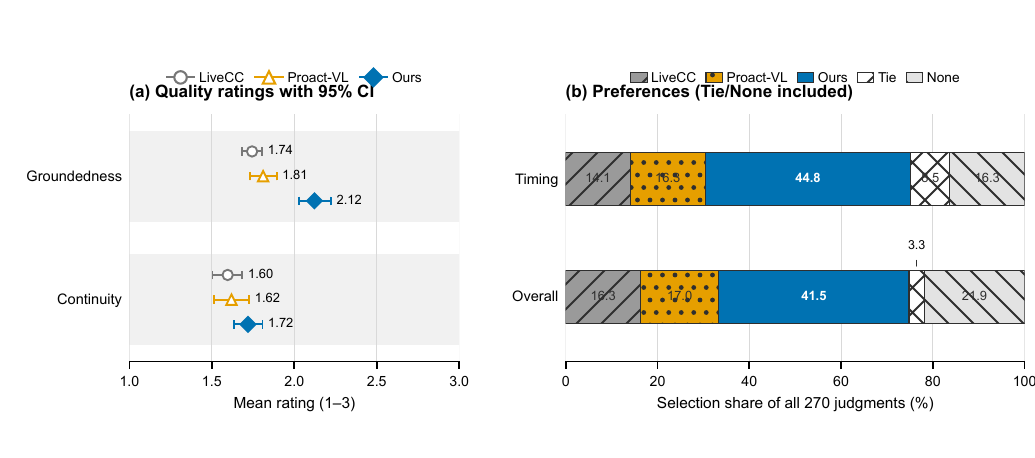}
\caption{Anonymous end-to-end human evaluation on 90 unique 20-second windows from 30 rounds. Three raters viewed synchronized muted footage and anonymous timestamped commentary tracks. Left: mean Groundedness and Continuity ratings with 30-round-clustered 95\% bootstrap confidence intervals. Right: Timing and Overall selections over all 270 window--rater judgments, including Tie and None. Ours receives the highest Groundedness and preference shares; Continuity is directionally higher, but its paired differences are unresolved.}
\label{fig:supp_e2e_human}
\end{figure*}

Three annotators follow the same written instructions and complete a nine-window pilot before formal evaluation; pilot responses are excluded. The study compares LiveCC, Proact-VL, and Our Planner on 90 unique 20-second windows from 30 prespecified BoxComm-Eval rounds, with three non-overlapping windows per round. Candidate windows require at least two utterance onsets from every system. Selection uses onset counts, normalized round position, and punch-event density, and balances early/middle/late position, low/medium/high event density, and three relative output-density strata (30 windows each). Each system retains its natural within-window timing, including pauses and silence. This active-window sample assesses comparative commentary quality when all systems speak; the full-match automatic protocol separately evaluates silence allocation and output density over complete broadcasts.

Raters view synchronized muted footage and three anonymous timestamped commentary tracks, including every within-window utterance and at most two preceding utterances per system as unscored context. Each rater receives an independently shuffled window order and independently balanced A/B/C mapping; over the 90 windows plus nine hidden repeats, every system appears exactly 33 times in each display position. For each method, raters assign a 1--3 \textbf{Groundedness} score for support from the visible action and context, and a 1--3 \textbf{Continuity} score for completeness, local coherence, and avoidance of disruptive repetition. They also select the best \textbf{Timing} and \textbf{Overall} output, with Tie and None options. Clarity and interestingness are not scored as separate items.

The nine hidden repeats are used only for quality control and do not enter method comparison. Confidence intervals use 2,000 bootstrap replicates clustered by the 30 source rounds, preserving the paired method and rater observations within each sampled round.

Figure~\ref{fig:supp_e2e_human} summarizes the window-level ratings and preferences. Over all 270 judgments, Ours receives 44.81\% of Timing and 41.48\% of Overall selections; the remaining responses include another method, Tie, or None. Among judgments selecting a single method, the corresponding shares are 121/203 (59.6\%) and 112/202 (55.4\%), as reported in the main paper. Paired comparisons in Table~\ref{tab:supp_e2e_deltas} show that the Groundedness and preference advantages remain positive under round-clustered resampling. Continuity is directionally higher, but both paired confidence intervals include zero; we therefore do not interpret it as a statistically established advantage.

\begin{center}
\centering
\small
\begin{tabular}{llcc}
\toprule
\textbf{Baseline} & \textbf{Measure} & \textbf{Difference} & \textbf{95\% CI} \\
\midrule
LiveCC & Groundedness & +0.378 & [0.256, 0.515] \\
LiveCC & Continuity & +0.122 & [$-$0.022, 0.270] \\
LiveCC & Timing & +0.307 & [0.204, 0.419] \\
LiveCC & Overall & +0.252 & [0.141, 0.367] \\
Proact-VL & Groundedness & +0.311 & [0.178, 0.456] \\
Proact-VL & Continuity & +0.100 & [$-$0.033, 0.241] \\
Proact-VL & Timing & +0.285 & [0.178, 0.396] \\
Proact-VL & Overall & +0.244 & [0.137, 0.359] \\
\bottomrule
\end{tabular}
\captionof{table}{Paired Ours-minus-baseline differences with round-clustered 95\% bootstrap confidence intervals. Timing and Overall differences are in selection-share units.}
\label{tab:supp_e2e_deltas}
\end{center}

\begin{figure*}[t]
\centering
\includegraphics[width=\textwidth]{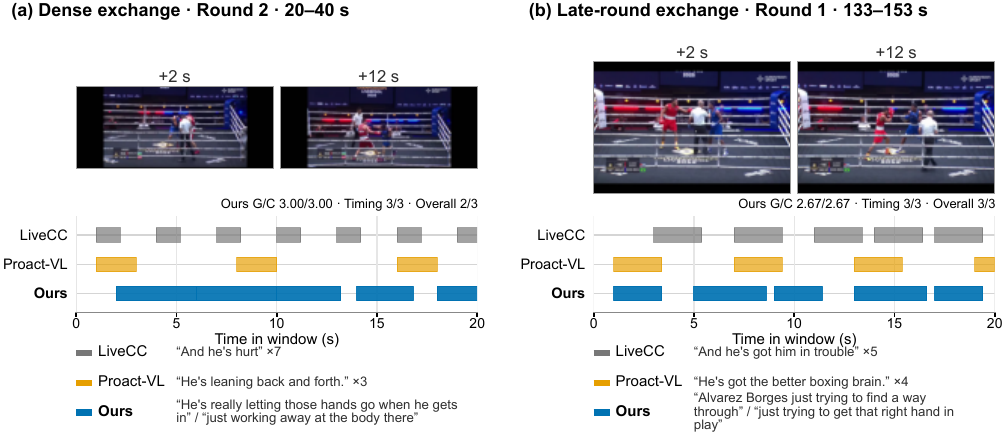}
\caption{Streaming examples from the formal human evaluation. Bars show every audible interval in each anonymous 20-second window, and quotes are representative verbatim outputs. Ours follows each exchange with varied action and tactical remarks, while the baselines repeat generic statements. Aggregate results over all 90 windows appear in Figure~\ref{fig:supp_e2e_human}.}
\label{fig:supp_e2e_qualitative}
\end{figure*}

Figure~\ref{fig:supp_e2e_qualitative} illustrates the two dimensions measured by the study. The Planner places utterances throughout active exchanges, while EIC-Gen changes the content as the action develops. This combination yields commentary that is better aligned with the visible bout and less dominated by a single repeated phrase in these examples.

\paragraph{Study reliability.}
Hidden-repeat exact consistency is 66.7\% for Groundedness, 79.0\% for Continuity, 66.7\% for Timing, and 70.4\% for Overall. Method comparisons use paired, round-clustered intervals: they support the Groundedness, Timing, and Overall differences, while the Continuity intervals include zero.

\section{Dataset Scope and Release}

\subsection{Contextual Commentary Audit}
A model-assisted, author-reviewed audit covers 100 category-qualified items across the 40 evaluation matches. We retained 80 items from an initial sample and replaced 12 items classified as another primary function, five ambiguous items, and three fragments. Replacement followed a fixed within-match candidate order and preserved each match's quota. The audit found 47 items about fighters or competition background, 24 about scores, time, or rules, nine about audiences or broadcasting, and 20 conversational responses or transitions. Nine of the 47 background items had support in the four-second video window. Across all 100 items, 35 lacked support in both the window and the supplied professional history. These include background facts as well as missing local broadcast cues. Two conflicts and seven unresolved source cases were kept separately. These sample counts motivate retrieval for background facts and richer broadcasting inputs for local context. Banter is labeled by its function: a fighter anecdote is Contextual, while an informal explanation of tactics is Tactical. The broad functions also allow finer sport-specific labels. Tactical commentary, for example, could distinguish punch selection within an exchange from strategy across a bout.

\begin{center}
\centering
\scriptsize
\begin{tabular*}{\linewidth}{@{\extracolsep{\fill}}p{0.27\linewidth}p{0.29\linewidth}p{0.34\linewidth}@{}}
\toprule
\textbf{Component} & \textbf{Release form} & \textbf{Reproducibility role} \\
\midrule
Code and checkpoints & Source, fixed rubrics, author weights & Training and evaluation \\
Commentary annotations & Text, times, and categories & Reference task construction \\
Punch timelines & Derived event records & Direct EIC-Gen/Planner inputs \\
Broadcasts & Source references; rights apply & Media acquisition and alignment \\
Proprietary punch subset & Not redistributed & Not required for downstream use of released timelines \\
\bottomrule
\end{tabular*}
\captionof{table}{Current and planned release components.}
\label{tab:supp_release}
\end{center}

BoxComm is drawn from one professional boxing championship and a common broadcast-source family. Its chronological split places later matches in the 40-match evaluation set, which covers all 20 weight classes across the women's and men's divisions. Planner development uses an internal 20-match subset of BoxComm-Train, followed by fixed comparisons on BoxComm-Eval. The released split establishes a controlled within-source benchmark and support future cross-broadcast challenge sets.

Category-conditioned generation uses professional preceding commentary, while streaming evaluation uses each system's generated history. Contextual generation is closed-book: no external fighter profile or retrieval corpus is supplied. The resulting tasks support controlled study of local visual, event, and discourse context, and provide a basis for future retrieval-grounded and long-horizon extensions.

The public Hugging Face dataset repository provides source URLs, crop metadata, Train/Eval manifests, commentary annotations, and derived punch-event archives. Original broadcasts remain with their rights holders. To reconstruct the benchmark input from an authorized source copy, the planned package will document the broadcast time base, exact crop start and end with their endpoint conventions, local timestamp offsets, and media identity checks. Changes to source availability or encoding can affect reconstruction.

We plan to release reconstruction and evaluation code, model configurations and author-trained weights, prompts and parsers, per-example predictions and judgments, and file checksums. Third-party weights follow their upstream licenses. The released punch timelines support downstream experiments without access to the proprietary punch subset used to train the extractor. A hosted evaluation on Hugging Face is planned to reduce reliance on independently acquiring broadcast copies.

\FloatBarrier
\section{Prompts and Model-Call Contracts}
\label{sec:supp_prompts}

This section lists the fixed instructions used for transcript construction, dataset annotation, commentary generation, content evaluation, and streaming-output classification. Angle-bracketed fields denote instance-specific payloads. We plan to release executable templates, parsers, model identifiers, outputs, and call manifests.

\begin{center}
\tiny
\begin{tabular*}{\linewidth}{@{\extracolsep{\fill}}p{0.21\linewidth}p{0.16\linewidth}p{0.14\linewidth}p{0.29\linewidth}@{}}
\toprule
\textbf{Stage} & \textbf{Model} & \textbf{Sampling} & \textbf{Returned object} \\
\midrule
ASR correction / split & GPT-5.2 & temperature 0 & JSON list of sentence strings \\
Pause-boundary check & GPT-5.2 & temperature 0 & split decision and local classes \\
Dataset category label & GPT-5.2 & temperature 0 & unchanged records plus class 1/2/3 \\
EIC-Gen train / test & listed VLM & teacher forcing / greedy test & one 6--20-word sentence \\
Content Consistency & GPT-5.2 & temperature 0 & indexed binary decisions \\
Streaming category & GPT-5.2 & temperature 0 & indexed class 1/2/3 decisions \\
\bottomrule
\end{tabular*}
\captionof{table}{Inventory of model-call roles used in the reported benchmark pipeline.}
\label{tab:supp_prompt_inventory}
\end{center}

\subsection{Transcript Construction}

WhisperX supplies the words and their timestamps. GPT-5.2 is asked only to correct clear domain homophones and recover sentence boundaries while retaining lexical content and order.

\paragraph{ASR correction and semantic segmentation.}
\begin{lstlisting}[style=boxcommprompt]
[SYSTEM]
You are a text segmentation tool for boxing commentary. Split ASR chunks into self-contained sentences using the ACTION-ANALYSIS split rule and others below. Do NOT paraphrase or rewrite content; preserve original wording as much as possible. Also silently correct obvious boxing ASR homophones (e.g. 'job'->'jab' when meaning a punch, 'fainting'->'feinting'). Return ONLY a JSON array of strings.

[USER]
Fighter name spelling reference - Blue corner: <BLUE_NAME>, Red corner: <RED_NAME>. When the ASR has a word that sounds like any of these names, correct it to match.

Split this boxing commentary chunk into self-contained sentences.

RULES:
1) ACTION-ANALYSIS SPLIT (highest priority): When a sentence combines an immediate visual action with a tactical intent or analysis - often joined by a comma, "trying to", "looking to", "as he/she tries", "in order to" - SPLIT them into two sentences.
   e.g. "Good jab, she's getting some success early." -> ["Good jab.", "She's getting some success early."]
   e.g. "Another straight shot from Scott, trying to set his feet, trying to send the big shots." -> ["Another straight shot from Scott.", "Trying to set his feet, trying to send the big shots."]
2) VISUAL-SHIFT SPLIT: Split when moving from background / story to live action.
3) PRESERVE INTEGRITY: Keep multiple phrases that describe ONE specific combo or action together.
4) NO DANGLING FRAGMENTS: Every resulting sentence must be a complete thought. Don't end on "but" / "and" / "as".
5) Preserve wording: keep original lexical content and order; do not paraphrase.
6) ASR CORRECTION: Fix obvious boxing homophones ("job"->"jab" as punch, "fainting"->"feinting").
7) If already a single clean sentence with one semantic type, return as-is.

Chunk: <ASR_CHUNK>
\end{lstlisting}

\paragraph{Middle-pause boundary validation.}
\begin{lstlisting}[style=boxcommprompt]
[SYSTEM]
You are assisting sentence-boundary validation for boxing commentary ASR. Given ONE candidate long-pause boundary inside a sentence, decide whether it should be split. Pause duration is a strong cue: if gap >= 4.0s, usually split unless grammar is tightly bound; if 2.0s<=gap<4.0s, split when the right side begins a new proposition/discourse move. Do NOT split fixed collocations or tightly bound phrases. Balance precision and recall: allow split when readability and self-contained meaning improve. If split=true, assign class for each side: class 1 play-by-play, class 2 tactical, class 3 contextual. Return ONLY valid JSON with keys: split (bool), reason (short), left_class (1|2|3|null), right_class (1|2|3|null).

[USER JSON]
{"file": <SOURCE_FILE>, "segment_index": <SEGMENT_INDEX>,
 "current_text": <TEXT>, "current_class": <CLASS>,
 "gap_seconds": <SECONDS>, "boundary_hint": <BOUNDARY>,
 "left_local_words": <LEFT>, "right_local_words": <RIGHT>,
 "prev2_text": <PREV2>, "prev_text": <PREV>,
 "next_text": <NEXT>, "next2_text": <NEXT2>,
 "output_schema": {"split": "boolean", "reason": "short string",
  "left_class": "1|2|3 or null", "right_class": "1|2|3 or null"}}
\end{lstlisting}

If either segmentation response is invalid, the deterministic fallback does not introduce a new boundary: the original chunk is retained for the first call, and the candidate pause remains unsplit for the second. This makes malformed responses conservative with respect to GPT-created segmentation. The pipeline additionally applies timestamp-based pause splitting and unambiguous 
\texttt{fainting}$\rightarrow$\texttt{feinting} normalization deterministically. The mid-gap call uses only the three short class names for child relabeling; it does not reuse the detailed ontology below, and an invalid child class retains its parent segment's class.

\subsection{Offline Category Annotation}

The labeler receives up to 20 marked targets, with the preceding five and following three sentences used only to disambiguate the offline label. This neighboring text is part of dataset construction, not category-conditioned or streaming model input.

\begin{lstlisting}[style=boxcommprompt]
[SYSTEM]
You are a boxing commentary analyst. Classify each sentence into exactly one class based on what it conveys and whether it can be seen on screen. Use surrounding context only to disambiguate - classify ONLY the explicitly marked sentences. Return ONLY a valid JSON array, no markdown, no explanation.

[USER: ONTOLOGY]
CLASS 1 - Play-by-Play Commentary
A specific, observable action or ring event happening RIGHT NOW, verifiable from a 1-3 s clip.

Qualifies as Class 1:
- Named punch / exchange / defence / movement: "jab lands", "steps back", "slips the jab", "one-two combination", "backhand to the body again"
- Current position / state: "back on the ropes", "clinch", "heads clash", "off balance"
- Ring events: "referee warns", "bell", "standing count", "score 10-9"
- Direct exclamation reacting to a specific visible action: "Oh! Great shot!", "Brilliant!", "Lovely right hand!"
- "trying to" / "targeting" + named physical action or body part: "targeting the body", "trying to land the hook", "going to the body" -> Class 1
- Demonstrative + specific named action: "Look at that hook!", "That right hand just got through"

Does NOT qualify as Class 1 - use Class 2 instead:
- "trying to" + abstract goal: "trying to find his range", "trying to force the pace"
- Comparative / quality judgment: "that's a bit better", "a bright start", "growing into the round"
- Habitual pattern ("feinting at times", "has been X-ing") - pattern over multiple moments, not one action

CLASS 2 - Tactical Commentary
Requires inference, pattern recognition, or professional boxing knowledge.
- Intent / tactics: "trying to find his range", "looking to dictate", "playing for the counter"
- Comparative judgments: "that's a bit better", "starting to find his range"
- Pattern / trend / momentum: "controlling the distance", "has been the busier fighter", "tiring as the rounds go on"
- General quality (no named trigger): "a bright start", "impressive display", "growing into the round"
- Prescriptions / predictions: "needs to cut the ring off", "should go to the body"
- Summary / consequence: "he's won that exchange", "second time X has been warned"

CLASS 3 - Contextual Commentary
Information that frames and contextualizes the action beyond the immediate live exchange. These lines provide narrative context for understanding fighters, event setting, and storyline.
- Fighter bios / career / physique: name, nationality, record, titles ("gold medal", "world champion"), physical build / strength
- Event info: venue, weight division, championship name, round structure
- History / culture: past results, previous meetings, career arc; generic nationality style claims
- Cross-match references (past tense OR present perfect): "I saw a knockdown too", "I have seen him put someone to sleep", "in their previous meeting" -> Class 3
- Broadcast filler: commentary intros, crowd atmosphere, score-card disputes ("I strongly disagree with this"), timing calls ("we've had 30 seconds"), transitions ("here we go", "underway"), replay intros ("let's look at the action from round 1"), short vague affirmations / cross-talk ("Yeah.", "Absolutely.", "Exactly.")

---
Classify ONLY the [CLASSIFY THESE] sentences below. Context lines are for reference only - do NOT include them in output.

Return a JSON array, one object per sentence (same order):
  {"text": <unchanged>, "start_frame": <unchanged>, "end_frame": <unchanged>, "start_time": <unchanged>, "end_time": <unchanged>, "class": <1|2|3>}
RETURN ONLY THE JSON ARRAY.

### Context - sentences BEFORE this batch (reference only, do NOT classify):
  [<START_FRAME>-<END_FRAME>] "<UP TO FIVE SENTENCES>"
### Context - sentences AFTER this batch (reference only, do NOT classify):
  [<START_FRAME>-<END_FRAME>] "<UP TO THREE SENTENCES>"
### [CLASSIFY THESE]:
<TARGET RECORDS>
\end{lstlisting}

The examples operationalize the main Play-by-Play/Tactical boundary: a named current action is Play-by-Play, whereas an inferred goal or multi-moment pattern is Tactical. Operationally, Contextual also covers non-action, non-tactical broadcast framing, including background information, transitions, crowd or replay framing, timing cues, and brief backchannels. A malformed category response fails the corresponding file instead of assigning a default class.

\subsection{Category-Conditioned Commentary Generation}

The same instruction format is used for generator training and inference. The target class is always provided; the history and event lines are included only in conditions that use those inputs. A video object covering the current clip is attached separately to the user message.

\begin{lstlisting}[style=boxcommprompt]
[SYSTEM]
You are a live boxing commentator.

[USER]
Previous commentary: <UP TO EIGHT PREVIOUS SENTENCES OR (none)>
Previous punch events (context only): <ORDERED PAST EVENTS>
Current time: <TIME>s

Event usage rules for Play-by-Play:
- Previous punch events are soft context and may be noisy.
- Prioritize what is visible in the current clip over previous events.
- Use event details only when consistent with current visuals.
- Do NOT output repetitive generic templates; be specific to this moment.

Event usage rules for Tactical or Contextual:
- Previous punch events are optional context only.
- Keep focus on current clip and target class style.

Class definitions:
- Class 1 (Play-by-Play): Describe observable actions/events happening now.
- Class 2 (Tactical): Analyze strategy, positioning, and intent.
- Class 3 (Contextual): Provide background, history, or broader match context.
Target class for this sample: <TARGET CLASS>.
Generate one natural <TARGET CLASS> commentary sentence for the current moment. Use 6-20 words. Output text only.
\end{lstlisting}

Only the event guidance matching the target class is inserted. The no-Event ablation removes both the event line and event guidance; other inputs and decoding remain fixed. At test time Qwen generation is greedy (temperature 0, top-$p$ 1.0, 64-token limit, repetition penalty 1.05). The 6--20-word request is a generation-style instruction rather than a target filter or hard truncation.

\subsection{Automatic Evaluation Calls}

\paragraph{Content Consistency rubric.}
The metric uses a semantic-overlap test because a single professional reference is not an exhaustive description of the moment. It accepts paraphrases and omissions of minor detail, while rejecting a different action, topic, or contradiction of the core meaning. It is therefore a lenient reference-alignment measure, not an independent verification of every factual attribute or a general hallucination metric.

\begin{lstlisting}[style=boxcommprompt]
[SYSTEM]
You are a strict JSON-only evaluator.

[USER]
You are an evaluator for sports commentary semantic overlap.
For each pair, output score=1 if prediction and reference are roughly about the same action/topic.
Be lenient to paraphrase, generic wording, missing minor details, and extra non-conflicting details.
Output score=0 only when prediction clearly talks about a different action/topic or contradicts core meaning.
Return JSON ONLY as an array of objects: [{"i":0,"score":0|1}, ...].
Do not add explanation.

Pairs:
[{"i": <ID>, "reference": <REFERENCE>, "prediction": <PREDICTION>}, ...]
\end{lstlisting}

GPT-5.2 receives identity-blind indexed pairs. Responses are schema-validated. We retain the evaluator outputs for the planned release.

\paragraph{Streaming-output Category classifier (short v1).}
This call maps realized generated sentences to the three narrated states used by t-IoU and four-state JS. It does not assess content correctness, and it is deliberately shorter than the context-rich offline annotation prompt.

\begin{lstlisting}[style=boxcommprompt]
[SYSTEM]
You are a boxing commentary analyst. Classify each sentence into exactly one class 1/2/3 and return JSON only.

[USER]
CLASS 1 - Play-by-Play Commentary
Immediate, observable ring actions right now.

CLASS 2 - Tactical Commentary
Strategy, intent, momentum, pattern, prediction, or analysis.

CLASS 3 - Contextual Commentary
Background/framing/storyline beyond immediate exchange.

Classify ONLY [TARGET] and keep order unchanged.
Return JSON ONLY as an array of objects:
[{"i":0,"class":1|2|3}, ...]
No explanation.

### [TARGET]:
[{"i": <ID>, "text": <GENERATED SENTENCE>}, ...]
\end{lstlisting}

All systems use this classifier after their outputs have been converted to sentences. The classifier sees neither method identity nor reference categories. Single-word backchannels are assigned Contextual by the same deterministic ontology rule for every system. Responses are schema-validated before scoring. The Planner and punch extractor are trained prediction modules and use no language-model prompt.